\documentclass[10pt,twocolumn,letterpaper]{article}

\usepackage[pagenumbers]{iccv} 

\definecolor{TableBlue}{rgb}{0.17,0.49,0.75}
\definecolor{Cerulean}{rgb}{0,0,0.95}
\definecolor{LimeGreen}{rgb}{0.15,0.65,0.15}
\definecolor{RoyalBlue}{rgb}{0.25,0.41,0.88}
\definecolor{Rose}{rgb}{1.0, 0.15, 0.21}
\definecolor{Orange}{rgb}{1.0, 0.5, 0.0}
\definecolor{Gray}{gray}{0.6}
\definecolor{Black}{gray}{0.0}
\definecolor{Purple}{rgb}{0.77,0.12,0.64}

\usepackage{acro}
\usepackage{tikz}
\usepackage{pifont}
\usepackage{multirow}
\usepackage{makecell}
\usepackage{subcaption}
\usepackage[table]{xcolor}

\def\secref#1{Sec.~\ref{#1}}

\def\tabref#1{Tab.~\ref{#1}}
\def\eqref#1{Eq.~(\ref{#1})}

\DeclareAcronym{VLM}{
  short = VLM,
  long  = vision-language model
}

\DeclareAcronym{VLMs}{
  short = VLMs,
  long  = vision-language models
}

\definecolor{iccvblue}{rgb}{0.21,0.49,0.74}
\usepackage[pagebackref,breaklinks,colorlinks,allcolors=iccvblue]{hyperref}

\def\paperID{****} 
\def\confName{ICCV}
\def\confYear{2025}

\title{AGO: Adaptive Grounding for Open World 3D Occupancy Prediction}

\author{
Peizheng~Li$^{1,2}$, Shuxiao~Ding$^{1,4}$, You~Zhou$^{1}$, Qingwen~Zhang$^{5}$, Onat~Inak$^{1,6}$, \\
Larissa~Triess$^{1}$, Niklas~Hanselmann$^{1,2,3}$, Marius~Cordts$^{1}$, Andreas~Zell$^{2}$ \\
$^{1}$Mercedes-Benz AG, Sindelfingen, $^{2}$University of Tübingen, $^{3}$Tübingen AI Center,\\
$^{4}$University of Bonn, $^{5}$RPL, KTH Royal Institute of Technology, $^{6}$TU Berlin\\
{\tt\small \{firstname.lastname\}@mercedes-benz.com},
{\tt\small qingwen@kth.se},
{\tt\small andreas.zell@uni-tuebingen.de}
}

\begin{document}

\twocolumn[{%
    \renewcommand\twocolumn[1][]{#1}%
    \maketitle
    \vspace{-5pt}
    \centering
    \includegraphics[width=0.9\textwidth]{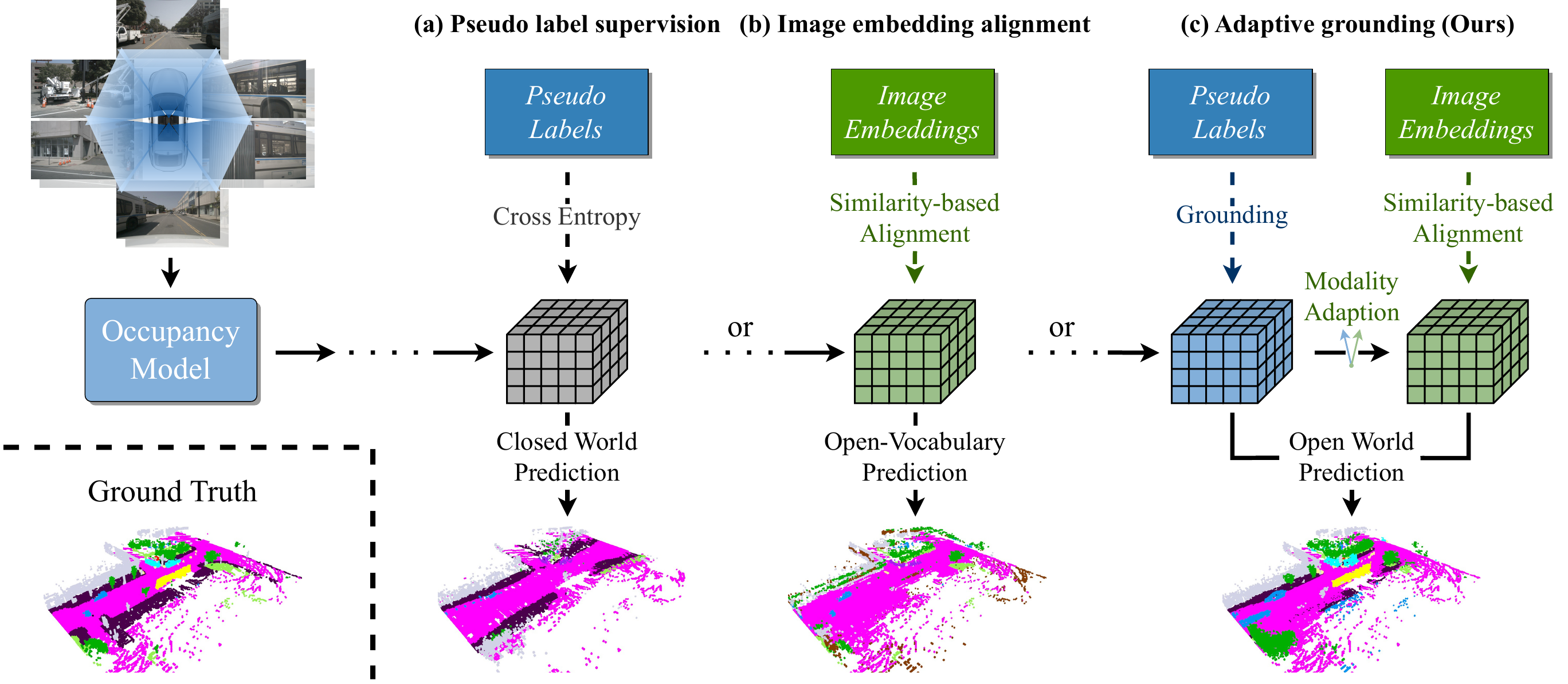}
    \captionof{figure}{
    \textbf{Open-world 3D semantic occupancy prediction.}
    (a) Supervision based on pseudo labels with fixed classes cannot predict novel categories, such as the ``\textcolor[RGB]{0, 255, 255}{construction vehicle}'' and ``\textcolor[RGB]{0, 175, 0}{vegetation}''.
    (b) Similarity-based alignment suffers from significant mismatches due to issues like modality discrepancy, leading to confusions between \eg ``\textcolor[RGB]{75, 0, 75}{sidewalk}'' and ``\textcolor[RGB]{255, 0, 255}{driveable surface}''.
    (c) Our proposed Adaptive Grounding flexibly accommodates both known and unknown objects, achieving more precise open world occupancy prediction.
    }
    \label{fig:overview}
    \vspace{10pt}
}]

\begin{abstract}
%
%
Open-world 3D semantic occupancy prediction aims to generate a voxelized 3D representation from sensor inputs while recognizing both known and unknown objects. 
Transferring open-vocabulary knowledge from vision-language models (VLMs) offers a promising direction but remains challenging. 
%
%
However, methods based on VLM-derived 2D pseudo-labels with traditional supervision are limited by a predefined label space and lack general prediction capabilities.
Direct alignment with pretrained image embeddings, on the other hand, often fails to achieve reliable performance because of inconsistent image and text representations in VLMs.
%
%
To address these challenges, we propose AGO, a novel 3D occupancy prediction framework with adaptive grounding to handle diverse open-world scenarios.
AGO first encodes surrounding images and class prompts into 3D and text embeddings, respectively, leveraging similarity-based grounding training with 3D pseudo-labels. 
Additionally, a modality adapter maps 3D embeddings into a space aligned with VLM-derived image embeddings, reducing modality gaps.
%
%
Experiments on Occ3D-nuScenes show that AGO improves unknown object prediction in zero-shot and few-shot transfer while achieving state-of-the-art closed-world self-supervised performance, surpassing prior methods by 4.09 mIoU.
Code is available at: https://github.com/EdwardLeeLPZ/AGO.
\vspace{-10pt}
\end{abstract}    
\section{Introduction}
\label{sec:intro}

Robust 3D scene representations are essential to autonomous driving systems, as they inherently encode rich geometric and semantic details.
One effective way for achieving such representations is 3D semantic occupancy prediction, which constructs voxelized structures with semantic labels from sensor inputs.
Although supervised approaches~\cite{huang2023tri, tian2024occ3d, wei2023surroundocc, zhang2023occformer, li2023voxformer, boeder2024occflownet, zhao2024hybridocc, shi2024occupancy} deliver strong performance, their reliance on extensive 3D annotations within closed label spaces restricts both training scalability and generalization to unlabeled unknown objects.

\begin{figure}[t]
    \centering
    \includegraphics[width=\linewidth]{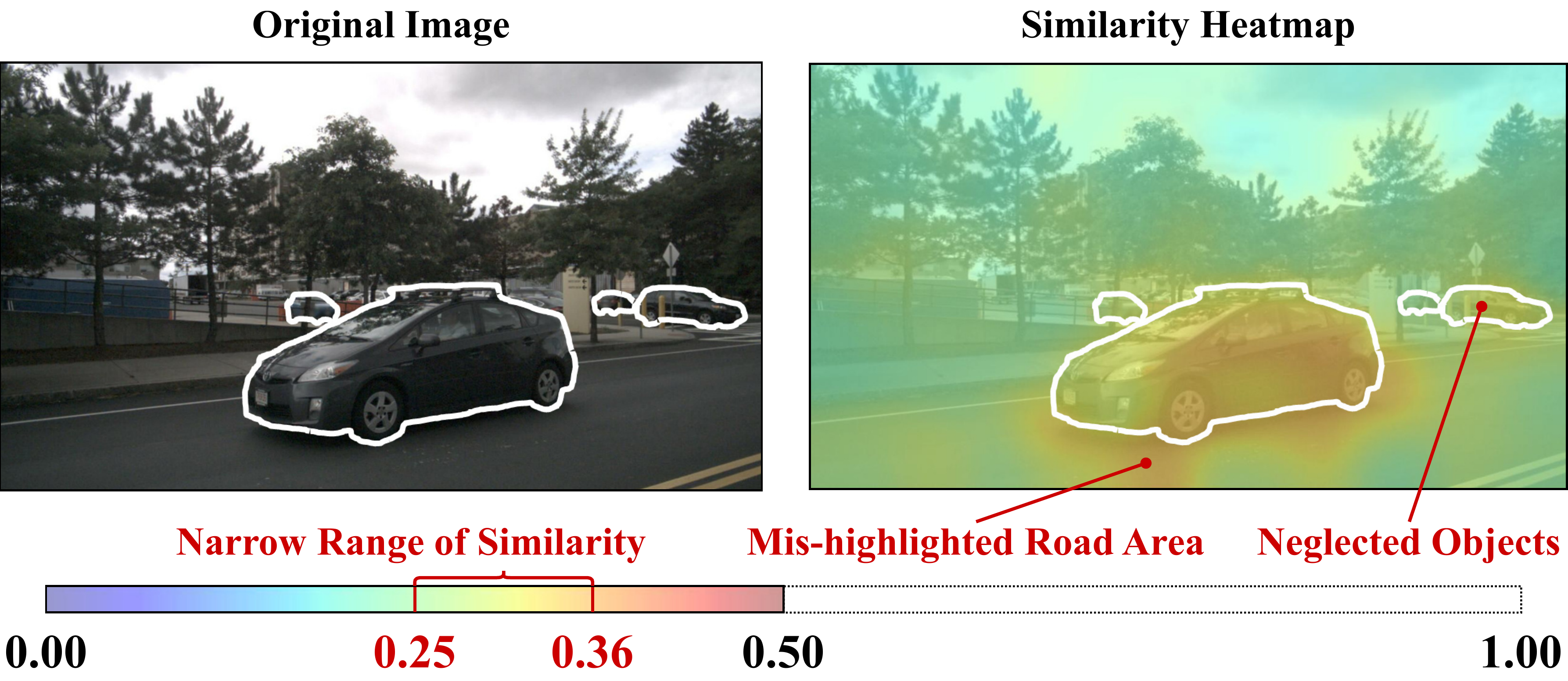}
    \vspace{-15pt}
    \caption{
    \textbf{Modality gaps between image and text embeddings in pre-trained \ac{VLMs}.}
    The similarity heatmap generated using the image and prompt ``car'' only covers a small range of about 0.1 (with a peak of only 0.36), while the low resolution of the embeddings also results in spatial misalignment and semantic ambiguity.
    }
    \label{fig:modality_gap}
    \vspace{-15pt}
\end{figure}

Recent advances incorporate open-vocabulary knowledge through \ac{VLMs}~\cite{zhang2024vision} to address these limitations.
As shown in Fig~\ref{fig:overview}.a and \ref{fig:overview}.b, several works generate 2D semantic pseudo-labels for 3D self-supervised training~\cite{zhang2023occnerf, huang2024selfocc, zheng2024veon, gan2024gaussianoccfullyselfsupervisedefficient}, while others directly align 3D features with VLM-derived 2D embeddings via knowledge distillation~\cite{tan2023ovo, vobecky2024pop, boeder2024langocc}.
However, most \ac{VLMs} are pre-trained on 2D vision-text data and lack explicit 3D geometric reasoning, which poses challenges in transferring their knowledge effectively.
Additionally, limited closed label spaces and inherent modality gaps (\Cref{fig:modality_gap}) can impair both semantic fidelity and geometric completeness.

To address these issues, we propose \textbf{A}daptive \textbf{G}rounding for 3D \textbf{O}ccupancy Prediction (\textbf{AGO}) illustrated in~\Cref{fig:overview}.c.
Our method constructs voxel representations for both closed-world and open-world objects and applies targeted alignment strategies to bridge the gap between the text and 2D image embeddings of the VLM.
During training, VLM-generated pseudo-labels enable grounding between 3D voxel and text embeddings, enhanced by negative sampling of additional noise text prompts. 
A modality adapter further maps 3D embeddings into a space aligned with VLM-derived image embeddings, facilitating knowledge transfer beyond predefined labels. 
During inference, an open-world identifier selects the most reliable embeddings based on information entropy for the final occupancy prediction.  

Experimental results show that AGO outperforms the state-of-the-art self-supervised methods on the Occ3D dataset~\cite{tian2024occ3d}. 
To further assess its open-world effectiveness, we introduce a benchmark consisting of 3 stages: pre-training on a limited label space, zero-shot evaluation on expanded classes, and few-shot fine-tuning, all without manual annotations.
Our proposed AGO maintains strong performance on known objects while achieving a clear advantage in predicting unknown objects, outperforming existing closed-world self-supervised methods.

The contributions of this paper are as follows: 
\begin{itemize}
    \item 
    We introduce AGO, a novel open-world occupancy prediction framework that adaptively distills knowledge from pretrained \ac{VLMs} into 3D perception for autonomous driving.
    \item We introduce grounding training with random text noise as negative samples to enhance the discriminative power of the 3D representation.
    \item We design a modality adapter to mitigate the differences between text and 2D image embeddings and an open world identifier to adaptively select features for known and unknown objects during inference in the open world.
    \item Our method achieves state-of-the-art performance on the closed-world self-supervised Occ3D-nuScenes benchmark and significantly outperforms existing methods in open-world occupancy prediction.
\end{itemize}

\begin{figure*}[htbp]
    \centering
    \includegraphics[width=0.92\textwidth]{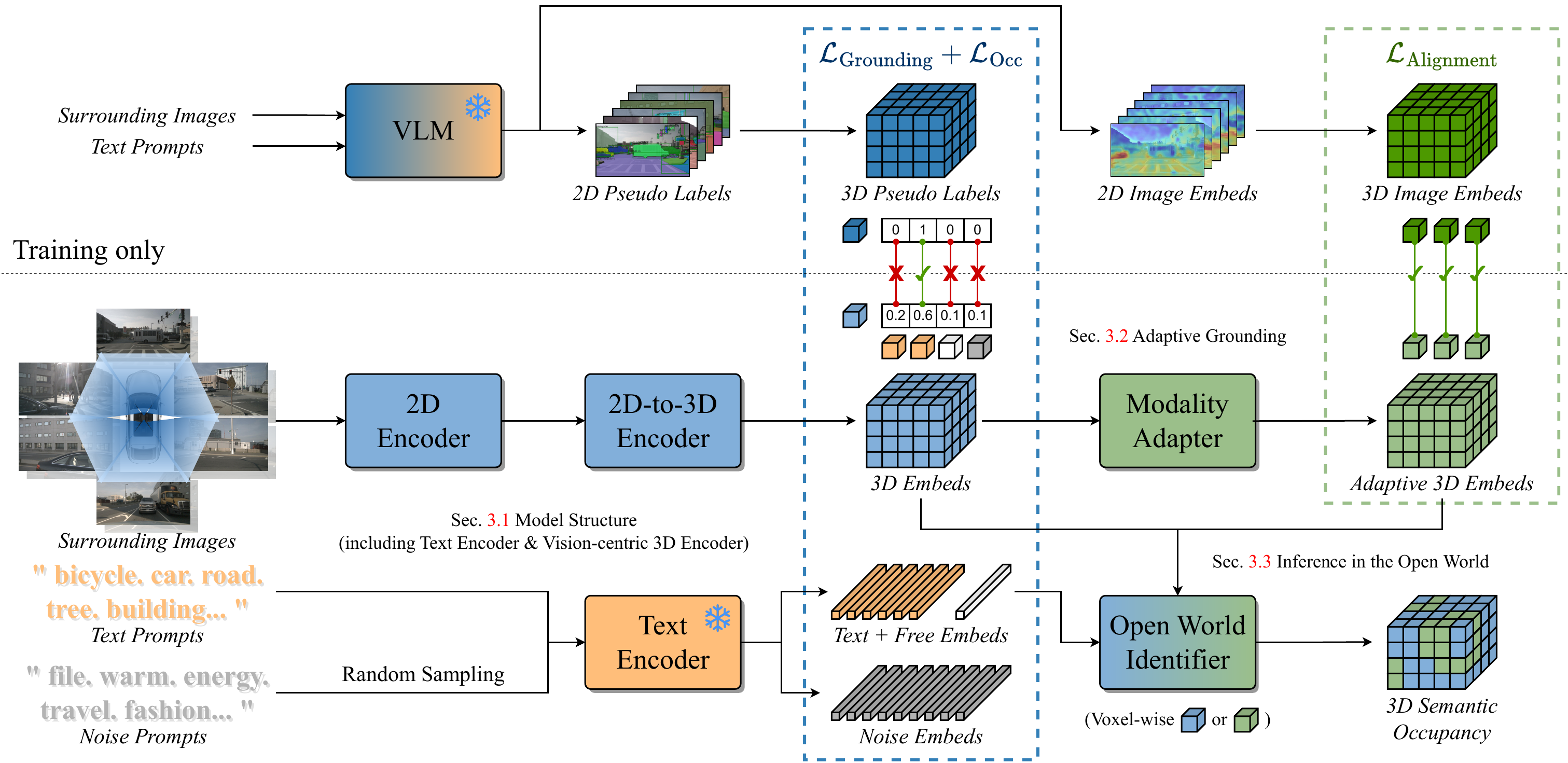}
    \vspace{-5pt}
    \caption{
    \textbf{AGO architecture.} 
    The upper part illustrates the generation process of 3D pseudo-labels and image embeddings based on pre-trained \ac{VLMs} during training.
    The lower part depicts the main architecture of our AGO framework, which comprises a frozen pre-trained text encoder, a vision-centric 3D encoder, a modality adapter, and an open-world identifier.
    The detailed illustration in the middle showcases our training paradigm, which consists of a noise-augmented grounding training and an adaptive image embedding alignment.
    }
    \label{fig:arc}
    \vspace{-10pt}
\end{figure*}
\section{Related Work}
\label{sec:related_work}

\noindent \textbf{3D scene understanding}
To accurately capture 3D environmental information, a common approach is to encode LiDAR point clouds into a global representation for various downstream tasks~\cite{zhou2018voxelnet, lang2019pointpillars, choy20194d, yin2021center, roldao2020lmscnet, xia2023scpnet, zhang2024deflow, zhang2025seflow, pillarseg, semanticvoxel}.
Recently, vision-centric approaches have gained increasing attention, extracting image features from perspective views~\cite{wang2022detr3d, liu2022petr, ding2025tqd} and lifting them into a BEV~\cite{philion2020lift, huang2021bevdet, li2022bevformer, zhou2022cross, jiang2023polarformer, li2023powerbev} or 3D voxel representation~\cite{cao2022monoscene, huang2023tri, tong2023scene, zhang2023occformer, li2023voxformer, wei2023surroundocc, ma2024cam4docc, zhao2024hybridocc, boeder2024occflownet, lin2024teocc, wang2024panoocc, yan2024renderworld}.
Among them, MonoScene~\cite{cao2022monoscene} first attempts to extract monocular 2D image features into 3D voxels using a line-of-sight projection, paving the way for future works.
Inspired by BEV-based models~\cite{li2022bevformer, zhou2022cross, jiang2023polarformer}, 
TPVFormer~\cite{huang2023tri} introduced a tri-perspective view representation, complementing BEV with two additional perpendicular planes for richer spatial encoding.
Subsequent methods refine prediction accuracy by improving feature representation. OccFormer~\cite{zhang2023occformer} enhances long-range and fine-grained feature interactions through global-local path decoupling, while VoxFormer~\cite{li2023voxformer} improves spatial consistency with a coarse-to-fine strategy.  
However, these approaches heavily rely on large-scale 3D annotations.
The high cost of manual labeling restricts their ability to adapt and scale to various, constantly evolving open-world driving situations.

\noindent \textbf{Multi-modal representation learning}
Multi-modal representation learning has expanded the scope of visual tasks.
Building on early works \cite{lu2019vilbert, li2019visualbert, tan2019lxmert, chen2020uniter}, CLIP~\cite{radford2021learning} introduced a dual-stream architecture with contrastive learning to align textual and visual modalities. 
Trained on 400 million image-text pairs, CLIP demonstrated strong zero-shot capabilities and inspired further research into large-scale cross-modal pre-training. 
As visual tasks became more complex, GLIP~\cite{li2022grounded} unified object detection and phrase grounding by aligning visual regions with text descriptors.
Grounding DINO~\cite{liu2023grounding} extended this approach with a transformer-based deformable attention mechanism.
MaskCLIP+~\cite{zhou2022extract} leveraged pre-trained CLIP features for pixel-level tasks, achieving zero-shot semantic segmentation through pseudo-labeling. 
More recently, Grounded SAM~\cite{ren2024grounded} combined SAM~\cite{kirillov2023segment} with Grounding DINO to enable fine-grained segmentation using arbitrary text inputs, showing strong potential for open-world applications.
Despite their advances, these models are not designed for 3D reasoning and require further adaptation for autonomous driving.

\noindent \textbf{Open-vocabulary occupancy prediction}
Recently, many works have explored integrating multi-modal representations into 3D scene understanding to extend knowledge boundaries with their open-vocabulary capabilities.
One common paradigm~\cite{cao2023scenerf, zhang2023occnerf, huang2024selfocc, liu2024let, abualhanud2024self, zheng2024veon, gan2024gaussianoccfullyselfsupervisedefficient, wang2025uniocc, zhou2025autoocc} is using 2D pseudo-labels generated by pre-training multimodal models as supervision.
OccNeRF~\cite{zhang2023occnerf} constructs unbounded parameterized 3D occupancy fields and then renders these fields into depth maps, supervised through multi-frame photometric consistency.
SelfOcc~\cite{huang2024selfocc} follows the same paradigm and proposes a MVS-embedded strategy to improve depth optimization.
But the above methods are limited by the closed label space and do not have the ability of open-world prediction.
VEON~\cite{zheng2024veon} further improves scene understanding by integrating pre-trained depth estimation~\cite{ranftl2020towards} and \ac{VLM}~\cite{radford2021learning} through delicate structure design and two-stage training.
Alternatively, some methods~\cite{vobecky2024pop, boeder2024langocc, jiang2024gausstr} directly align 3D or reconstructed 2D features with multi-modal pre-trained embeddings.
POP-3D~\cite{vobecky2024pop} performs pixel-level alignment of 3D features and pre-trained CLIP~\cite{radford2021learning} image embeddings based on the LiDAR point cloud.
GaussTR~\cite{jiang2024gausstr}, on the other hand, constructs a sparse 3D Gaussian representation and splats it in the image plane for 2D alignment, reducing the dependence on LiDAR.
However, these methods achieved suboptimal performance due to the modality gaps within pre-trained models, low resolution, and incomplete scenes.
In summary, developing a 3D occupancy prediction framework that balances strong performance with open-world generalization remains an urgent challenge.
\section{Method}
\label{sec:method}
3D semantic occupancy prediction aims to convert sensor inputs into a voxelized semantic representation of the environment.
The vision-centric model takes $N_\text{cam}$ surrounding camera images $\mathbf{I} = \{ \mathbf{I}_n \}_{n = 1}^{N_\text{cam}} $ from a single timestamp as inputs and outputs semantic voxels $\mathbf{O} \in \mathbb{C}^{H \times W \times D}$, where $H$, $W$ and $D$ represent the spatial resolution of the 3D voxel space, and $\mathbb{C}$ is the label space containing $N_\text{cls}$ classes.

As illustrated in~\Cref{fig:arc}, our proposed AGO consists of four modules. The \textit{text encoder} converts label and noise prompts into text embeddings. The \textit{vision-centric 3D encoder} extracts 2D features from the input images and lifts them into 3D voxel embeddings. The \textit{modality adapter} projects these embeddings into a semantic space aligned with pretrained \ac{VLM} image embeddings. Finally, the \textit{open world identifier} selects the most appropriate features for occupancy prediction in the open-world scenarios.

\subsection{Model Structure}
\label{subsec:model_structure}
\noindent \textbf{Text encoder}
To extend a fixed, closed label space into a flexible, open semantic space, we take the labels as textual inputs and employ a text encoder to extract their semantic information.
Specifically, unlike the full sentences used in traditional 2D visual grounding, we decompose the given category labels into predefined finer-grained subcategory phrases serving text prompts.
These prompts are then tokenized and further encoded into text embeddings by a pre-trained language model, \eg BERT~\cite{devlin2018bert} in MaskCLIP+~\cite{zhou2022extract}.
Besides, we randomly add noise prompts by selecting phrases from a general dictionary. The resulting noise embeddings serve as negative samples during grounding training, thereby enhancing the discrimination capability of the model for closed-world objects.
The process described above can be summarized as:
\begin{equation}
    \mathbf{F}_T = \mathcal{E}_T(\mathbf{T}), \quad \mathbf{F}_{\text{noise}} = \mathcal{E}_T(\mathbf{T_{\text{noise}}}), 
    \label{eq:text_encoder}
\end{equation}
where $\mathbf{F}_T \in \mathbb{R}^{N_\text{text} \times C}$ and $\mathbf{F}_{\text{noise}} \in \mathbb{R}^{N_\text{noise} \times C}$. 
$N_\text{text}$ and $N_\text{noise}$ are the number of tokens of text prompts and noise prompts, which are roughly the same.
$C$ is the dimensionality of text embeddings.
We freeze the text encoder throughout the training process, thus preserving its open-vocabulary performance.
Compared to traditional supervised learning, a clear advantage is that the label space does not have to remain consistent between training and inference.

\noindent \textbf{Vision-centric 3D encoder}
Our framework follows the architecture of TPVFormer~\cite{huang2023tri} to construct an efficient 3D representation.
First, the surrounding images are fed into an image backbone~$\mathcal{E}_I$ to obtain 2D perspective features $\mathbf{F}_I^{\text{2D}} = \mathcal{E}_I(\mathbf{I})$.
Next, a 3D encoder lifts these~$\mathbf{F}_I^{\text{2D}}$ to mutually orthogonal tri-perspective views $\mathbf{F}_I^{TPV} = [\mathbf{F}_I^{HW}, \mathbf{F}_I^{DH}, \mathbf{F}_I^{WD}] = \mathcal{E}_{\text{3D}}(\mathbf{F}_I^{\text{2D}})$.
Finally, they are expanded and superimposed to form the full 3D embeddings $\mathbf{F}_I^{\text{3D}} \in \mathbb{R}^{H \times W \times D \times C}$, where $C$ denotes their feature dimensionality, identical to above text embeddings.
This process can be formulated as:
\begin{equation}
    \mathbf{F}_I^{\text{3D}} = \mathcal{E}_{\text{3D}}(\mathcal{E}_I(\mathbf{I})).
    \label{eq:2D_to_3D_encoder}
\end{equation}

\subsection{Adaptive Grounding}
\label{subsec:modality_adapter}
To achieve precise closed-world occupancy prediction while also enabling open-world transfer, we propose adaptive grounding that decouples training for known and unknown categories through grounding training and adaptive alignment, respectively.

\noindent \textbf{Closed-world grounding training}
For known categories, we construct 3D voxel pseudo-labels from 2D semantic masks generated by a pre-trained \ac{VLM}~\cite{ren2024grounded}.
Specifically, we define the label space according to common autonomous driving scenarios~\cite{vobecky2024pop, zheng2024veon, boeder2024langocc} and use the label phrases as text prompts to generate corresponding 2D semantic masks of the surrounding images.
These masks are then transformed into 3D occupancy pseudo-labels~$\mathbf{O}_{\text{pl}} \in {\mathbb{C}}^{H \times W \times D}$ via the geometric correspondence between LiDAR point clouds and camera pixels.
By incoporating the post-processing of multi-frame aggregation, point cloud ray casting, and semantic voting, this process produces a more comprehensive 3D semantic scene than what single-frame 2D supervision can achieve, thereby enhancing closed-world self-supervised performance. 

Traditional classifiers trained on pseudo-labels can only distinguish known categories and require retraining to recognize unlabeled objects.
Instead, we adopt grounding training by directly comparing the 3D voxel features \(\mathbf{F}_I^{\text{3D}}\) with text embeddings \(\mathbf{F}_T\) and noise embeddings \(\mathbf{F}_{\text{noise}}\).
In order to represent free (unoccupied) voxels, we also introduce a learnable embedding \(\mathbf{F}_\text{free}\) with the same dimension \(C\) as \(\mathbf{F}_T\).
These embeddings are concatenated into  
$\mathbf{F}_{T+\text{noise}+\text{free}} = [\mathbf{F}_T; \mathbf{F}_\text{noise}; \mathbf{F}_\text{free}]$.
We then compute the voxel-wise similarity score by taking the dot product:
\begin{equation}
  \mathbf{S}^{\text{3D}} = \mathbf{F}_I^{\text{3D}} \cdot \mathbf{F}_{T+\text{noise}+\text{free}}^\top.
  \label{eq:grounding_alignment_score}
\end{equation}
The resulting \(\mathbf{S}^{\text{3D}} \in \mathbb{R}^{H \times W \times D \times (N_\text{token}+N_\text{noise}+1)}\) acts as generalized logits, allowing us to apply standard occupancy prediction losses used in traditional supervised training.

We follow TPVFormer~\cite{huang2023tri}, using a combination of the cross-entropy loss and the Lovász-softmax loss~\cite{berman2018lovasz} for grounding training:
\begin{align}
    \mathcal{L}_{\text{Grounding}} &= \mathcal{L}_{\text{CE}}(\mathbf{S}_{\text{sem}}^{\text{3D}}, \mathbf{O}_{\text{pl}}) + \mathcal{L}_{\text{Lovász}}(\mathbf{S}_{\text{sem}}^{\text{3D}}, \mathbf{O}_{\text{pl}}), \;\text{with} \notag \\
    \mathbf{S}_{\text{sem}, k}^{\text{3D}} &= \frac{1}{|\mathcal{T}_k|} \sum_{i \in \mathcal{T}_k} \mathbf{S}_i^{\text{3D}},
    \label{eq:self_supervised_loss}
\end{align}
where \(\mathcal{T}_k\) denotes the set of token indices for category \(k\) and \(|\mathcal{T}_k|\) is the number of tokens in the set. 
We average the scores \(\{\mathbf{S}_i^{\text{3D}}\}\) from tokens belonging to the same category to form the semantic logits for both text and noise inputs.

Furthermore, we incorporate an occupancy loss to balance the occupied and free space:
\begin{align}
    \mathcal{L}_{\text{Occ}} &= \mathcal{L}_{\text{CE}}(\mathbf{S}_{\text{binary}}^{\text{3D}}, \mathbf{O}_{\text{binary}}), \; \text{with} \notag\\
    \mathbf{S}_{\text{binary}}^{\text{3D}} &= [\max(\mathbf{S}_{\text{sem}}^{\text{3D}} \setminus \mathbf{S}_{\text{sem}, \,\text{noise}+\text{free}}^{\text{3D}}); \;\mathbf{S}_{\text{sem}, \,\text{free}}^{\text{3D}}], \notag \\
    \mathbf{O}_{\text{binary}} &= \begin{cases} 0 & \text{if } \mathbf{O}_{\text{pl}} = \text{free} \\ 1 & \text{otherwise} \end{cases}
    \label{eq:occ_loss}
\end{align}
  
\noindent \textbf{Open-world adaptive alignment}
While grounding is an effective closed-world training paradigm, it cannot recognize unknown objects beyond fixed pseudo-labels.
In contrast, VLMs benefit from large-scale pretraining and produce image embeddings with richer semantics that implicitly represent unlabeled objects in autonomous driving scenes.
By using similarity-based alignment, these embeddings can be distilled into 3D representations, expanding their semantic scope.
However, as shown in~\Cref{sec:intro} and \Cref{fig:modality_gap}, alignment alone yields limited prediction performance.
A more promising approach is to combine similarity-based alignment with the grounding paradigm described above.

One straightforward implementation is to apply both loss functions simultaneously to the same 3D embedding.
However, since the two alignment objectives target different modalities—text and vision—naively adding their losses can lead to modality conflicts during training, ultimately degrading performance.
This is also supported by the experimental comparison in~\Cref{tab:alignment_ablation}.

To address these issues, we propose an adaptive alignment mechanism that assists grounding training to solve the unknown object prediction. 
We first map the 3D embeddings $\mathbf{F}_I^{\text{3D}}$ into a new semantic space as $\tilde{\mathbf{F}}_I^{\text{3D}} \in \mathbb{R}^{H \times W \times D \times C}$.
They are then aligned with the image embeddings of a pre-trained \ac{VLM}, \eg MaskCLIP+~\cite{zhou2022extract}, distilling the semantic-rich information contained in the image embeddings into 3D.
Specifically, the 2D image embeddings~$\mathbf{E}_I^{\text{2D}} \in \mathbb{R}^{H_I \times W_I \times C}$ are projected into the 3D voxels~$\mathbf{E}_I^{\text{3D}} \in \mathbb{R}^{H \times W \times D \times C}$ following the same geometric projection used for pseudo-label generation.
Here, we employ a lightweight MLP as the modality adapter ($\mathcal{MA}$):
\begin{equation}
    \tilde{\mathbf{F}}_I^{\text{3D}} = \mathcal{MA}(\mathbf{F}_I^{\text{3D}}).
    \label{eq:modality_adapter}
\end{equation}

For the alignment loss, we adopt cosine similarity. It computes the loss on non-empty voxels visible in the camera views, thereby avoiding mismatches due to occlusion.
\begin{equation}
    \mathcal{L}_{\text{Alignment}} = \frac{1}{|\mathcal{V_{\text{visible}}}|} \sum_{v \in \mathcal{V_{\text{visible}}}} \left( 1 - \cos(\tilde{\mathbf{F}}_{I, v}^{\text{3D}}, \mathbf{E}_{I, v}^{\text{3D}}) \right),
    \label{eq:alignment_loss}
\end{equation}
where $\mathcal{V}_{\text{visible}}$ denotes the set of visible non-empty voxels. Finally,
the overall loss function is formulated as:
\begin{equation}
    \mathcal{L}_{\text{Total}} = \mathcal{L}_{\text{Grounding}} + \mathcal{L}_{\text{Occ}} + \mathcal{L}_{\text{Alignment}}.
    \label{eq:total_loss}
\end{equation}

\subsection{Inference in the Open World}
\label{subsec:open_world_identifier} 
We propose an open world identifier for balanced decision-making between grounding and adaptive alignment in occupancy prediction. 
For each voxel $v \in \mathcal{V}$, it compares the similarity obtained from both 3D embedding and adaptive 3D embedding to determine which one to use for the final prediction.
Specifically, we compute the dot products ($\mathbf{S}_{\text{sem}, v}^{\text{3D}}$ and $\tilde{\mathbf{S}}_{\text{sem}, v}^{\text{3D}}$) between query text embeddings and the 3D embeddings as well as adaptive 3D embeddings according to the formula~Eqs.~(\ref{eq:grounding_alignment_score}, \ref{eq:self_supervised_loss}).
They are then passed through the softmax function and become the confidence scores ($\mathbf{P}_v$ and $\tilde{\mathbf{P}}_v$) for the voxel belonging to a particular category.

Based on statistical analysis in~\Cref{fig:ave_entropy} and experimental verification in~\Cref{tab:ow_strategy_ablation}, we use minimum information entropy as the criterion:
By comparing the entropy of $\mathbf{P}_v$ and $\tilde{\mathbf{P}}_v$ , we select the probability distribution with lower entropy:
\begin{align}
    \mathbf{P}_{\text{final}, v} &= \mathbf{c}_v \mathbf{P}_v + (1 - \mathbf{c}_v) \tilde{\mathbf{P}}_v, \;\text{with} \notag \\
    \mathbf{c}_v &= \mathbb{I} (\mathcal{H}(\mathbf{P}_v) \leq \mathcal{H}(\tilde{\mathbf{P}}_v))
    \label{eq:decision_making}
\end{align}
where  \(\mathbb{I}(\cdot)\) is the indicator function, $\mathbf{c}_v$ represents the criterion for voxel $v$ and $\mathcal{H}(\cdot)$ denotes the entropy.

\section{Experiments}
\label{sec:experiments}

\begin{table*}[htbp]
    \vspace{-5pt}
    \centering
    \resizebox{\textwidth}{!}{
    \begin{tabular}{l|c|ccccccccccccccccc|cc}
        \toprule
        Method 
        & Image Backbone
        & \rotatebox{90}{\parbox{1cm}{\raggedright\tikz\draw[fill={rgb,255: red,0; green,0; blue,0}, draw={rgb,255: red,0; green,0; blue,0}] (0,0) rectangle (0.2,0.2);~oth.}} 
        & \rotatebox{90}{\parbox{1cm}{\raggedright\tikz\draw[fill={rgb,255: red,255; green,120; blue,50}, draw={rgb,255: red,255; green,120; blue,50}] (0,0) rectangle (0.2,0.2);~bar.}} 
        & \rotatebox{90}{\parbox{1cm}{\raggedright\tikz\draw[fill={rgb,255: red,255; green,192; blue,203}, draw={rgb,255: red,255; green,192; blue,203}] (0,0) rectangle (0.2,0.2);~bic.}} 
        & \rotatebox{90}{\parbox{1cm}{\raggedright\tikz\draw[fill={rgb,255: red,255; green,255; blue,0}, draw={rgb,255: red,255; green,255; blue,0}] (0,0) rectangle (0.2,0.2);~bus}}
        & \rotatebox{90}{\parbox{1cm}{\raggedright\tikz\draw[fill={rgb,255: red,0; green,150; blue,245}, draw={rgb,255: red,0; green,150; blue,245}] (0,0) rectangle (0.2,0.2);~car}} 
        & \rotatebox{90}{\parbox{1cm}{\raggedright\tikz\draw[fill={rgb,255: red,0; green,255; blue,255}, draw={rgb,255: red,0; green,255; blue,255}] (0,0) rectangle (0.2,0.2);~c. v.}}
        & \rotatebox{90}{\parbox{1cm}{\raggedright\tikz\draw[fill={rgb,255: red,200; green,180; blue,0}, draw={rgb,255: red,200; green,180; blue,0}] (0,0) rectangle (0.2,0.2);~mot.}} 
        & \rotatebox{90}{\parbox{1cm}{\raggedright\tikz\draw[fill={rgb,255: red,255; green,0; blue,0}, draw={rgb,255: red,255; green,0; blue,0}] (0,0) rectangle (0.2,0.2);~ped.}}
        & \rotatebox{90}{\parbox{1cm}{\raggedright\tikz\draw[fill={rgb,255: red,255; green,240; blue,150}, draw={rgb,255: red,255; green,240; blue,150}] (0,0) rectangle (0.2,0.2);~t. c.}}
        & \rotatebox{90}{\parbox{1cm}{\raggedright\tikz\draw[fill={rgb,255: red,135; green,60; blue,0}, draw={rgb,255: red,135; green,60; blue,0}] (0,0) rectangle (0.2,0.2);~tra.}} 
        & \rotatebox{90}{\parbox{1cm}{\raggedright\tikz\draw[fill={rgb,255: red,160; green,32; blue,240}, draw={rgb,255: red,160; green,32; blue,240}] (0,0) rectangle (0.2,0.2);~tru.}}
        & \rotatebox{90}{\parbox{1cm}{\raggedright\tikz\draw[fill={rgb,255: red,255; green,0; blue,255}, draw={rgb,255: red,255; green,0; blue,255}] (0,0) rectangle (0.2,0.2);~d. s.}}
        & \rotatebox{90}{\parbox{1cm}{\raggedright\tikz\draw[fill={rgb,255: red,139; green,137; blue,137}, draw={rgb,255: red,139; green,137; blue,137}] (0,0) rectangle (0.2,0.2);~o. f.}}
        & \rotatebox{90}{\parbox{1cm}{\raggedright\tikz\draw[fill={rgb,255: red,75; green,0; blue,75}, draw={rgb,255: red,75; green,0; blue,75}] (0,0) rectangle (0.2,0.2);~sid.}}
        & \rotatebox{90}{\parbox{1cm}{\raggedright\tikz\draw[fill={rgb,255: red,150; green,240; blue,80}, draw={rgb,255: red,150; green,240; blue,80}] (0,0) rectangle (0.2,0.2);~ter.}}
        & \rotatebox{90}{\parbox{1cm}{\raggedright\tikz\draw[fill={rgb,255: red,230; green,230; blue,250}, draw={rgb,255: red,230; green,230; blue,250}] (0,0) rectangle (0.2,0.2);~man.}}
        & \rotatebox{90}{\parbox{1cm}{\raggedright\tikz\draw[fill={rgb,255: red,0; green,175; blue,0}, draw={rgb,255: red,0; green,175; blue,0}] (0,0) rectangle (0.2,0.2);~veg.}}
        & \rotatebox{90}{mIoU*}
        & \rotatebox{90}{mIoU}\\
        \midrule            
        SimpleOcc~\cite{gan2023simple} & ResNet-101 & 0.00 & 0.67 & 1.18 & 3.21 & 7.63 & 1.02 & 0.26 & 1.80 & 0.26 & 1.07 & 2.81 & 40.44 & 0.00 & 18.30 & 17.01 & 13.42 & 10.84 & 7.99 & 7.05 \\
        POP-3D$^{\dagger}$~\cite{vobecky2024pop} & ResNet-101 & 0.06 & 0.02 & 0.46 & 1.83 & 4.87 & 0.00 & 0.00 & 1.29 & 0.00 & 0.65 & 2.62 & \textbf{55.90} & 1.60 & 9.99 & 25.17 & 15.75 & 21.11 & 9.42 & 8.31 \\
        SelfOcc~\cite{huang2024selfocc} & ResNet-50 & 0.00 & 0.15 & 0.66 & 5.46 & 12.54 & 0.00 & 0.80 & 2.10 & 0.00 & 0.00 & 8.25 & 55.49 & 0.00 & \textbf{26.30} & 26.54 & 14.22 & 5.60 & 10.54 & 9.30 \\
        OccNeRF~\cite{zhang2023occnerf} & ResNet-101 & 0.00 & 0.83 & 0.82 & 5.13 & 12.49 & 3.50 & 0.23 & 3.10 & 1.84 & 0.52 & 3.90 & 52.62 & 0.00 & 20.81 & 24.75 & 18.45 & 13.19 & 10.81 & 9.53 \\
        GaussianOcc~\cite{gan2024gaussianoccfullyselfsupervisedefficient} & Swin & 0.00 & 1.79 & 5.82 & \textbf{14.58} & 13.55 & 1.30 & 2.82 & \textbf{7.95} & \textbf{\underline{9.76}} & 0.56 & 9.61 & 44.59 & 0.00 & 20.10 & 17.58 & 8.61 & 10.29 & 11.26 & 9.94 \\
        GaussTR~\cite{jiang2024gausstr} & VFMs & 0.00 & 2.09 & 5.22 & 14.07 & \textbf{20.34} & \textbf{5.70} & 7.08 & 5.12 & 3.93 & 0.92 & \textbf{13.36} & 39.44 & 0.00 & 15.68 & 22.89 & 21.17 & 21.87 & 13.26 & 11.70 \\
        LangOcc~\cite{boeder2024langocc} & ResNet-50 & 0.00 & 3.10 & \textbf{\underline{9.00}} & 6.30 & 14.20 & 0.40 & \textbf{10.80} & 6.20 & \textbf{9.00} & 3.80 & 10.70 & 43.70 & \textbf{2.23} & 9.50 & 26.40 & 19.60 & \textbf{\underline{26.40}} & 13.27 & 11.84 \\
        VEON~\cite{zheng2024veon} & ViT-L & \textbf{0.90} & \textbf{\underline{10.40}} & 6.20 & \textbf{\underline{17.70}} & 12.70 & \textbf{\underline{8.50}} & 7.60 & 6.50 & 5.50 & \textbf{8.20} & 11.80 & 54.50 & 0.40 & 25.50 & \textbf{30.20} & \textbf{25.40} & 25.40 & \textbf{17.07} & \textbf{15.14} \\
        \midrule
        AGO (ours) & ResNet-101 & \textbf{\underline{1.53}} & \textbf{6.75} & \textbf{6.43} & 14.00 & \textbf{\underline{22.82}} & 5.57 & \textbf{\underline{16.66}} & \textbf{\underline{13.20}} & 6.80 & \textbf{\underline{10.53}} & \textbf{\underline{15.89}} & \textbf{\underline{71.48}} & \textbf{\underline{4.48}} & \textbf{\underline{34.48}} & \textbf{\underline{41.37}} & \textbf{\underline{29.33}} & \textbf{25.66} & \textbf{\underline{21.39}} & \textbf{\underline{19.23}} \\
        \bottomrule
    \end{tabular}}
    \vspace{-5pt}
    \caption{
    \textbf{3D occupancy prediction performance under the self-supervised setting on the Occ3D-nuScenes~\cite{tian2024occ3d} dataset.} 
    The full names of all abbreviated categories can be found by color in~\Cref{fig:occ_vis}.
    ``VFMs'' stands for the ensemble of multiple vision foundation models.
    $^{\dagger}$ indicate values obtained from our retraining.
    We also calculate the result as ``mIoU*'' ignoring the ``other'' and ``other flat'' categories, while ``mIoU'' is the original result. 
    Results are highlighted in \textbf{\underline{bold \& underlined}} for the best performance and \textbf{bold} for the second-best performance.
    }
    \label{tab:selfsupervised_benchmark}
    \vspace{5pt}
\end{table*}

\begin{table*}[htbp]
    \vspace{-5pt}
    \centering
    \resizebox{\textwidth}{!}{
    \begin{tabular}{l|l|ccc|cc|c|ccccccc|ccccc|c|c}
        \toprule
        Training Stages
        & Method
        & \rotatebox{90}{\parbox{1cm}{\raggedright\tikz\draw[fill={rgb,255: red,255; green,0; blue,0}, draw={rgb,255: red,255; green,0; blue,0}] (0,0) rectangle (0.2,0.2);~ped.}}
        & \rotatebox{90}{\parbox{1cm}{\raggedright\tikz\draw[fill={rgb,255: red,255; green,0; blue,255}, draw={rgb,255: red,255; green,0; blue,255}] (0,0) rectangle (0.2,0.2);~d. s.}}
        & \rotatebox{90}{\parbox{1cm}{\raggedright\tikz\draw[fill={rgb,255: red,75; green,0; blue,75}, draw={rgb,255: red,75; green,0; blue,75}] (0,0) rectangle (0.2,0.2);~sid.}}
        & \rotatebox{90}{\parbox{1cm}{\raggedright\tikz\draw[fill={rgb,255: red,0; green,128; blue,128}, draw={rgb,255: red,0; green,128; blue,128}] (0,0) rectangle (0.2,0.2);~veh.}}
        & \rotatebox{90}{\parbox{1cm}{\raggedright\tikz\draw[fill={rgb,255: red,128; green,0; blue,128}, draw={rgb,255: red,128; green,0; blue,128}] (0,0) rectangle (0.2,0.2);~cyc.}}
        & \rotatebox{90}{k. mIoU}
        & \rotatebox{90}{\parbox{1cm}{\raggedright\tikz\draw[fill={rgb,255: red,0; green,150; blue,245}, draw={rgb,255: red,0; green,150; blue,245}] (0,0) rectangle (0.2,0.2);~car}}
        & \rotatebox{90}{\parbox{1cm}{\raggedright\tikz\draw[fill={rgb,255: red,255; green,255; blue,0}, draw={rgb,255: red,255; green,255; blue,0}] (0,0) rectangle (0.2,0.2);~bus}}
        & \rotatebox{90}{\parbox{1cm}{\raggedright\tikz\draw[fill={rgb,255: red,0; green,255; blue,255}, draw={rgb,255: red,0; green,255; blue,255}] (0,0) rectangle (0.2,0.2);~c. v.}}
        & \rotatebox{90}{\parbox{1cm}{\raggedright\tikz\draw[fill={rgb,255: red,135; green,60; blue,0}, draw={rgb,255: red,135; green,60; blue,0}] (0,0) rectangle (0.2,0.2);~tra.}} 
        & \rotatebox{90}{\parbox{1cm}{\raggedright\tikz\draw[fill={rgb,255: red,160; green,32; blue,240}, draw={rgb,255: red,160; green,32; blue,240}] (0,0) rectangle (0.2,0.2);~tru.}}
        & \rotatebox{90}{\parbox{1cm}{\raggedright\tikz\draw[fill={rgb,255: red,255; green,192; blue,203}, draw={rgb,255: red,255; green,192; blue,203}] (0,0) rectangle (0.2,0.2);~bic.}} 
        & \rotatebox{90}{\parbox{1cm}{\raggedright\tikz\draw[fill={rgb,255: red,200; green,180; blue,0}, draw={rgb,255: red,200; green,180; blue,0}] (0,0) rectangle (0.2,0.2);~mot.}} 
        & \rotatebox{90}{\parbox{1cm}{\raggedright\tikz\draw[fill={rgb,255: red,255; green,120; blue,50}, draw={rgb,255: red,255; green,120; blue,50}] (0,0) rectangle (0.2,0.2);~bar.}}  
        & \rotatebox{90}{\parbox{1cm}{\raggedright\tikz\draw[fill={rgb,255: red,255; green,240; blue,150}, draw={rgb,255: red,255; green,240; blue,150}] (0,0) rectangle (0.2,0.2);~t. c.}}
        & \rotatebox{90}{\parbox{1cm}{\raggedright\tikz\draw[fill={rgb,255: red,150; green,240; blue,80}, draw={rgb,255: red,150; green,240; blue,80}] (0,0) rectangle (0.2,0.2);~ter.}}
        & \rotatebox{90}{\parbox{1cm}{\raggedright\tikz\draw[fill={rgb,255: red,230; green,230; blue,250}, draw={rgb,255: red,230; green,230; blue,250}] (0,0) rectangle (0.2,0.2);~man.}}
        & \rotatebox{90}{\parbox{1cm}{\raggedright\tikz\draw[fill={rgb,255: red,0; green,175; blue,0}, draw={rgb,255: red,0; green,175; blue,0}] (0,0) rectangle (0.2,0.2);~veg.}}
        & \rotatebox{90}{u. mIoU}
        & \rotatebox{90}{mIoU} \\
        
        \midrule
        
        \multirow{4}{*}{\makecell[l]{Pretraining}} & POP-3D$^{\dagger}$~\cite{vobecky2024pop} & \cellcolor{green!25}0.00 & \cellcolor{green!25}58.77 & \cellcolor{green!25}13.80 & \cellcolor{cyan!25}8.27 & \cellcolor{cyan!25}1.10 & 16.39 & - & - & - & - & - & - & - & \cellcolor{red!25}0.00 & \cellcolor{red!25}0.00 & \cellcolor{red!25}3.95 & \cellcolor{red!25}\textbf{0.13} & \cellcolor{red!25}0.60 & 0.94 & 8.66 \\
        
        & SelfOcc$^{\dagger}$~\cite{huang2024selfocc} & \cellcolor{green!25}0.98 & \cellcolor{green!25}60.29 & \cellcolor{green!25}14.68 & \cellcolor{cyan!25}7.11 & \cellcolor{cyan!25}0.00 & 16.61 & - & - & - & - & - & - & - & \cellcolor{red!25}0.00 & \cellcolor{red!25}0.00 & \cellcolor{red!25}0.00 & \cellcolor{red!25}0.00 & \cellcolor{red!25}0.00 & 0.00 & 8.31 \\

        & GaussTR$^{\dagger}$~\cite{jiang2024gausstr} & \cellcolor{green!25}6.11 & \cellcolor{green!25}60.06 & \cellcolor{green!25}18.02 & \cellcolor{cyan!25}6.77 & \cellcolor{cyan!25}2.25 & 18.64 & - & - & - & - & - & - & - & \cellcolor{red!25}0.00 & \cellcolor{red!25}0.00 & \cellcolor{red!25}4.95 & \cellcolor{red!25}0.07 & \cellcolor{red!25}8.05 & 2.61 & 10.63 \\
        
        & AGO (ours) & \cellcolor{green!25}\textbf{7.82} & \cellcolor{green!25}\textbf{63.09} & \cellcolor{green!25}\textbf{25.53} & \cellcolor{cyan!25}\textbf{9.19} & \cellcolor{cyan!25}\textbf{5.03} & \textbf{22.13} & - & - & - & - & - & - & - & \cellcolor{red!25}0.00 & \cellcolor{red!25}0.00 & \cellcolor{red!25}\textbf{7.04} & \cellcolor{red!25}0.03 & \cellcolor{red!25}\textbf{10.88} & \textbf{3.59} & \textbf{12.86} \\
        
        \midrule
        
        \multirow{4}{*}{\makecell[l]{Zero-shot \\ Evaluation}} & POP-3D$^{\dagger}$~\cite{vobecky2024pop}  & \cellcolor{green!25}0.00 & \cellcolor{green!25}58.77 & \cellcolor{green!25}13.80 & - & - & 24.19 & \cellcolor{orange!25}6.72 & \cellcolor{orange!25}0.00 & \cellcolor{orange!25}0.00 & \cellcolor{orange!25}0.59 & \cellcolor{orange!25}4.34 & \cellcolor{orange!25}1.17 & \cellcolor{orange!25}\textbf{1.20} & \cellcolor{red!25}0.00 & \cellcolor{red!25}0.00 & \cellcolor{red!25}3.95 & \cellcolor{red!25}0.13 & \cellcolor{red!25}0.60 & 1.56 & 6.08\\
        
        & SelfOcc$^{\dagger}$~\cite{huang2024selfocc} & \cellcolor{green!25}0.98 & \cellcolor{green!25}60.29 & \cellcolor{green!25}14.68 & - & - & 25.23 & \cellcolor{orange!25}0.00 & \cellcolor{orange!25}0.00 & \cellcolor{orange!25}0.00 & \cellcolor{orange!25}0.00 & \cellcolor{orange!25}0.00 & \cellcolor{orange!25}0.00 & \cellcolor{orange!25}0.00 & \cellcolor{red!25}0.00 & \cellcolor{red!25}0.00 & \cellcolor{red!25}0.00 & \cellcolor{red!25}0.00 & \cellcolor{red!25}0.00 & 0.00 & 5.06\\

        & GaussTR$^{\dagger}$~\cite{jiang2024gausstr} & \cellcolor{green!25}6.11 & \cellcolor{green!25}60.06 & \cellcolor{green!25}18.02 & - & - & 28.06 & \cellcolor{orange!25}5.07 & \cellcolor{orange!25}\textbf{1.65} & \cellcolor{orange!25}0.00 & \cellcolor{orange!25}0.04 & \cellcolor{orange!25}1.84 & \cellcolor{orange!25}2.58 & \cellcolor{orange!25}0.27 & \cellcolor{red!25}0.00 & \cellcolor{red!25}0.00 & \cellcolor{red!25}4.95 & \cellcolor{red!25}0.07 & \cellcolor{red!25}8.05 & 2.04 & 7.25\\
        
        & AGO (ours) & \cellcolor{green!25}\textbf{7.82} & \cellcolor{green!25}\textbf{63.09} & \cellcolor{green!25}\textbf{25.53} & - & - & \textbf{32.15} & \cellcolor{orange!25}\textbf{7.67} & \cellcolor{orange!25}0.00 & \cellcolor{orange!25}0.00 & \cellcolor{orange!25}\textbf{1.33} & \cellcolor{orange!25}\textbf{6.50} & \cellcolor{orange!25}\textbf{4.50} & \cellcolor{orange!25}0.00 & \cellcolor{red!25}0.00 & \cellcolor{red!25}0.00 & \cellcolor{red!25}\textbf{7.04} & \cellcolor{red!25}0.03 & \cellcolor{red!25}\textbf{10.88} & \textbf{3.16} & \textbf{8.96}\\
        
        \midrule
        
        \multirow{4}{*}{\makecell[l]{Few-shot \\ Finetuning}} & POP-3D$^{\dagger}$~\cite{vobecky2024pop} & \cellcolor{green!25}0.00 & \cellcolor{green!25}44.90 & \cellcolor{green!25}12.79 & - & - & 19.23 & \cellcolor{orange!25}5.59 & \cellcolor{orange!25}0.03 & \cellcolor{orange!25}0.00 & \cellcolor{orange!25}0.29 & \cellcolor{orange!25}2.05 & \cellcolor{orange!25}1.26 & \cellcolor{orange!25}1.03 & \cellcolor{red!25}0.00 & \cellcolor{red!25}0.00 & \cellcolor{red!25}5.72 & \cellcolor{red!25}0.21 & \cellcolor{red!25}6.75 & 1.91 & 5.37\\
        
        & SelfOcc$^{\dagger}$~\cite{huang2024selfocc} & \cellcolor{green!25}7.85 & \cellcolor{green!25}65.65 & \cellcolor{green!25}25.29 & - & - & 32.93 & \cellcolor{orange!25}1.41 & \cellcolor{orange!25}0.00 & \cellcolor{orange!25}0.00 & \cellcolor{orange!25}0.01 & \cellcolor{orange!25}0.00 & \cellcolor{orange!25}0.00 & \cellcolor{orange!25}0.00 & \cellcolor{red!25}0.00 & \cellcolor{red!25}0.00 & \cellcolor{red!25}3.63 & \cellcolor{red!25}6.04 & \cellcolor{red!25}10.96 & 1.84 & 8.06\\

        & GaussTR$^{\dagger}$~\cite{jiang2024gausstr} & \cellcolor{green!25}7.84 & \cellcolor{green!25}66.36 & \cellcolor{green!25}25.55 & - & - & 33.25 & \cellcolor{orange!25}10.85 & \cellcolor{orange!25}1.58 & \cellcolor{orange!25}0.00 & \cellcolor{orange!25}0.00 & \cellcolor{orange!25}1.32 & \cellcolor{orange!25}1.42 & \cellcolor{orange!25}0.00 & \cellcolor{red!25}0.00 & \cellcolor{red!25}0.00 & \cellcolor{red!25}12.74 & \cellcolor{red!25}9.12 & \cellcolor{red!25}8.16 & 3.77 & 9.66\\
        
        & AGO (ours) & \cellcolor{green!25}\textbf{13.00} & \cellcolor{green!25}\textbf{71.54} & \cellcolor{green!25}\textbf{29.91} & - & - & \textbf{38.15} & \cellcolor{orange!25}\textbf{18.73} & \cellcolor{orange!25}\textbf{5.49} & \cellcolor{orange!25}0.00 & \cellcolor{orange!25}\textbf{0.41} & \cellcolor{orange!25}\textbf{2.16} & \cellcolor{orange!25}\textbf{3.72} & \cellcolor{orange!25}\textbf{2.22} & \cellcolor{red!25}\textbf{0.43} & \cellcolor{red!25}0.00 & \cellcolor{red!25}\textbf{29.63} & \cellcolor{red!25}\textbf{21.43} & \cellcolor{red!25}\textbf{17.73} & \textbf{8.50} & \textbf{14.43}\\
        \bottomrule
    \end{tabular}}
    \vspace{-5pt}
    \caption{
    \textbf{3D occupancy prediction performance under the open-world setting on the Occ3D-nuScenes~\cite{tian2024occ3d} dataset.}
    $^{\dagger}$ indicate values obtained from our retraining.
    The background color represents whether the category is known or unknown during the \textbf{pre-training stage}: \colorbox{green!25}{green} and \colorbox{cyan!25}{blue} indicate known IoU, \colorbox{orange!25}{orange} and \colorbox{red!25}{red} indicate unknown IoU.
    The \colorbox{orange!25}{orange} categories are the refined version of the \colorbox{cyan!25}{blue} categories.
    Results are highlighted in \textbf{bold} for the best performance.
    }
    \label{tab:open_world_benchmark}
    \vspace{-5pt}
\end{table*}

\subsection{Experimental Settings}
\label{subsec:experimental_settings}
\noindent \textbf{Dataset and metrics}
Our experiments are conducted on the Occ3D-nuScenes benchmark~\cite{tian2024occ3d}, which is derived from the nuScenes dataset~\cite{caesar2020nuscenes} that consists of 700/150/150 scenes for train/val/test splits, totaling 40,000 frames.
This benchmark uses nuScenes sensor data and each frame includes six-camera images and a 32-beam LiDAR point cloud. 
Occ3D-nuScenes provides voxelized 3D semantic occupancy labels with a spatial range of [-40m, 40m]$\times$[-40m, 40m]$\times$[-1m, 5.4m] and a resolution of 0.4m$\times$0.4m$\times$0.4m.
We evaluate performance using per-category IoU and the overall mIoU.

\noindent \textbf{Task settings}
We evaluate baselines and our method under two settings.
The \emph{\textbf{self-supervised task}} follows the Occ3D-nuScenes benchmark~\cite{tian2024occ3d} but only uses pseudo-labels during training.
In the \emph{\textbf{open-world task}}, we simulate real-world conditions by gradually expanding the label space. 
As shown in~\Cref{tab:open_world_benchmark}, the model is first pretrained with pseudo-labels with only five major categories (the first five columns in green and blue), with the last two being supercategories composed of multiple nuScenes classes.
It is then evaluated in zero-shot settings on the full Occ3D label space. 
By this time, the supercategories (blue) are split into their original classes (orange) and the remaining classes are added as fully unknown categories (red).
Afterwards, the model is fine-tuned based on a small number of samples to validate few-shot generalization ability of unknown classes.
Further details are provided in the supplementary material.

\noindent \textbf{Implementation details}
We use TPVFormer~\cite{huang2023tri} with ResNet-101~\cite{he2016deep} image backbone as the vision-centric 3D encoder.
It produces a voxel grid with the shape $200 \times 200 \times 16$.
Moreover, we adopt BERT~\cite{devlin2018bert} with MaskCLIP+~\cite{zhou2022extract} pre-trained weights as the text encoder.
For higher resolution, we utilize the MaskCLIP+~\cite{zhou2022extract} image features upsampled by FeatUp~\cite{fu2024featup} as 2D image embeddings.
For the modal adapters in AGO, we use a 2-layer MLP with a softplus activation.
Given its superior mask quality, we employ Grounded SAM~\cite{ren2024grounded} to generate 2D pseudo-labels.
During training, we utilize the AdamW~\cite{loshchilov2017decoupled} optimizer with a learning rate of $10^{-3}$ as well as the cosine learning rate scheduler, which starts with a linear warm-up from $10^{-5}$ for 500 steps and finishes at $10^{-6}$.
For self-supervised training and open-world pre-training tasks, we train our model for 24 epochs following the Occ3D~\cite{tian2024occ3d} setting, while for open-world fine-tuning tasks, we train our model for 12 epochs.
All experiments are conducted on 8 × NVIDIA A100 GPUs with a batch size of 1 per GPU.
Further details can be found in the supplementary material.

\subsection{Results and Analysis}
\label{subsec:results_and_analysis}

\begin{figure*}[htbp]
    \centering
    \includegraphics[width=\textwidth]{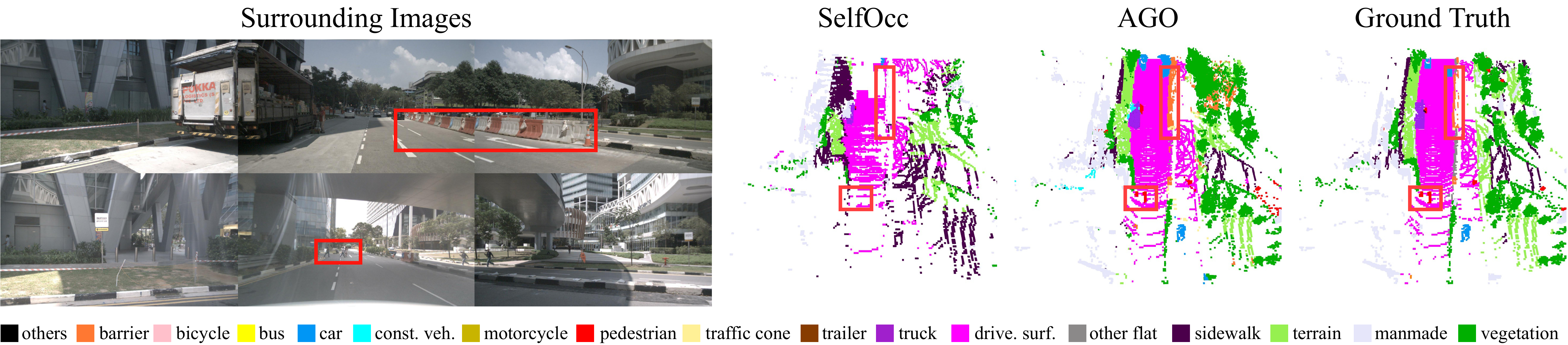}
    \vspace{-15pt}
    \caption{
    \textbf{Visualization of self-supervised 3D semantic occupancy prediction on the Occ3D-nuScenes occupancy benchmark.} 
    Our method demonstrates more detailed predictions for dynamic (``pedestrian'') and long-tailed (``barrier'') objects.
    }
    \label{fig:occ_vis}
\end{figure*}

\noindent \textbf{Closed-world self-supervised benchmark}
\Cref{tab:selfsupervised_benchmark} shows the prediction performance of AGO in the closed-world self-supervised task.
Our method demonstrates notable performance improvement, surpassing the previous best approach~\cite{zheng2024veon} by 4.09 mIoU.
Notably, AGO not only obtains significant gains in many static classes (\eg +15.58 in ``driveable surface'', +8.18 in ``sidewalk'', +11.17 in ``terrain''), but also achieves the state of the art in dynamic categories (\eg +2.48 in ``car'', +5.86 in ``motorcycle'', +5.25 in ``pedestrian'', +2.53 in ``truck'').
Unlike methods that rely on volume rendering or Gaussian splatting for 3D-to-2D projections, AGO directly optimizes 3D voxel features without requiring specific voxel settings~\cite{zhang2023occnerf}, additional learnable modules~\cite{boeder2024langocc}, or complex consistency losses~\cite{huang2024selfocc,gan2024gaussianoccfullyselfsupervisedefficient, jiang2024gausstr}.
Some approaches also directly integrate large pretrained foundational models~\cite{zheng2024veon, jiang2024gausstr}, leading to high computational costs (\eg, VEON~\cite{zheng2024veon} with 678.1M parameters). 
In contrast, AGO achieves the state-of-the-art performance without relying on delicate training setting or large-scale models (requiring only 62.5M parameters).
Additionally, we present the qualitative results.
As shown in~\Cref{fig:occ_vis}, our method can produce more accurate predictions for dynamic objects and long-tailed classes.
These results highlight the robustness and effectiveness of AGO in learning accurate 3D semantic occupancy representations in a closed-world setting.

\begin{figure*}[htbp]
    \centering
    \includegraphics[width=\textwidth]{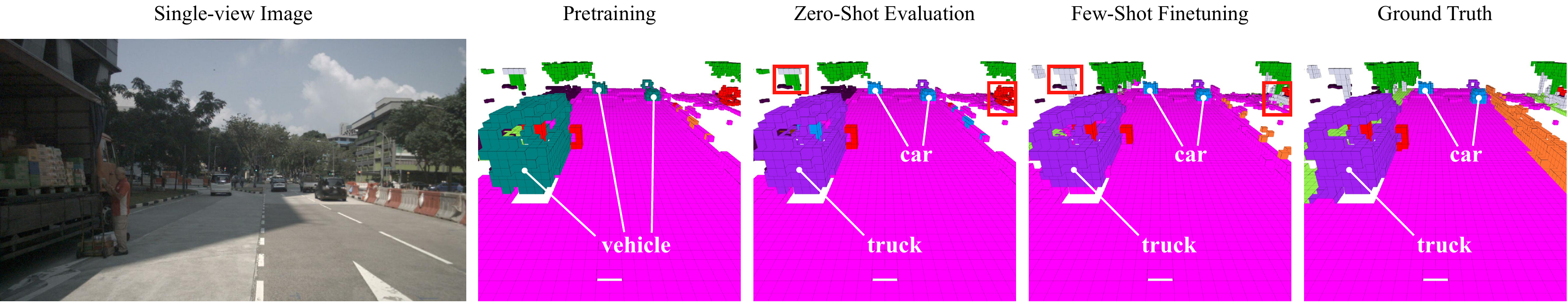}
    \vspace{-15pt}
    \caption{
    \textbf{Visualization of Open-world Zero-\&Few-Shot Transfer.}
    Our method can not only adapt to category changes from coarse to fine but also easily accommodate newly introduced, previously unknown categories with only a small amount of few-shot fine-tuning.
    }
    \label{fig:ow_occ_vis}
    \vspace{-5pt}
\end{figure*}

\noindent \textbf{Open-world evaluation and analysis}
We evaluate the generalization capability of AGO in the open-world task, as shown in~\Cref{tab:open_world_benchmark}.
To compare the open-world prediction performance of different training paradigms, we selected POP-3D~\cite{vobecky2024pop}, SelfOcc~\cite{huang2024selfocc} and GaussTR~\cite{jiang2024gausstr} as baselines.
POP-3D and GaussTR represent image embedding alignment methods, while SelfOcc relies on predefined closed-set pseudo-labels for supervision.
During pretraining (first four rows in~\Cref{tab:open_world_benchmark}), AGO maintains strong performance on known classes (22.13 mIoU) while also achieving non-trivial prediction for unknown categories such as 7.04 IoU for ``terrain'' and 10.88 IoU for ``vegetation''.
SelfOcc, restricted by its fixed label space, completely fails to predict unknown classes, yielding 0~mIoU across all of them.
POP-3D, despite leveraging a pre-trained VLM, struggles due to the lack of geometric and semantic cues in its pseudo-labels, reaching only 0.94 mIoU on average for unknown categories.
Thanks to the multiple integrated foundation models, GaussTR obtains better results but still performs lower than the 12.86 mIoU achieved by AGO.

When evaluated in a zero-shot setting with splitting supercategories into original classes, AGO outperforms POP-3D with IoU improvements of +0.95 (``car''), +2.16 (``truck''), and +3.33 (``bicycle'').
Similarly, GaussTR also struggles in these categories.
Notably, the IoU scores of ``truck'' (6.50) and ``bicycle'' (4.50) even surpass those of several methods in the closed-world benchmark~\Cref{tab:selfsupervised_benchmark}. 
Although the closed-world SelfOcc may capture semantic similarity, they are not trained on the new labels of supercategories and thus cannot handle the coarse-to-fine category transition.

Few-shot transfer aims to expand the knowledge of the model with minimal additional labeling.
We fine-tune all methods using a small amount of supercategories data. 
POP-3D, relying only on alignment, shows little improvement and overfits image embeddings, leading to a 4.95 mIoU drop in known classes (24.19~mIoU to 19.23~mIoU). 
SelfOcc also struggles in the open-world task, achieving only a +1.84 mIoU gain for unknown categories. 
While GaussTR benefits from fine-tuning, its alignment-based training limits improvement to just 2.41 mIoU. 
Our approach, in contrast, achieves consistent improvements in both known and unknown categories, with gains of +6.00 and +5.34 in mIoU, respectively, demonstrating its superior few-shot transfer capability in the open-world task.

As shown in~\Cref{fig:ow_occ_vis}, AGO generalizes effectively in zero-shot supercategory expansion (e.g., vehicle → car \& truck).
Few-shot fine-tuning further refines subcategory completeness while improving recognition of novel classes like ``manmade'' and ``vegetation''.
Overall, AGO balances open-world adaptability with strong closed-world performance. 

\subsection{Ablation Study}
\label{subsec:ablation_study}

\begin{table}
    \centering
    \footnotesize
    \resizebox{\linewidth}{!}{
        \begin{tabular}{c|cccc}
            \toprule
            \multirow{2}{*}{\makecell{Training \\ Paradigm}} & \multicolumn{4}{c}{mIoU} \\
            & Self. & O.W. Pre. & O.W. Z.S. & O.W. F.S. \\
            \midrule
            Align & 10.28 & 15.4\,/\,0.8\,/\,8.1 & 23.5\,/\,1.2\,/\,5.6 & 24.4\,/\,4.0\,/\,8.1\\
            Gro. & 19.08 & 20.6\,/\,0.0\,/\,10.3 & \textbf{33.0}\,/\,0.3\,/\,6.8 & \textbf{38.4}\,/\,3.0\,/\,10.1\\
            Gro. + Align & 18.89 & 18.3\,/\,2.2\,/\,10.2 & 29.3\,/\,1.4\,/\,7.0 & 37.3\,/\,5.7\,/\,12.0\\
            AGO & \textbf{19.23} & \textbf{22.1}\,/\,\textbf{3.6}\,/\,\textbf{12.9} & 32.2\,/\,\textbf{3.2}\,/\,\textbf{9.0} & 38.2\,/\,\textbf{8.5}\,/\,\textbf{14.4}\\
            \bottomrule
        \end{tabular}
    }
    \caption{
    \textbf{Ablation study of training paradigm.}
    ``Self.'' is short for Self-supervised and ``O.W.'' for Open World.
    ``Pre.'', ``Z.S.'' and ``F.S.'' represent three open-world stages: pretraining, zero-shot evaluation and few-shot finetuning.
    Each of their corresponding items is composed of three parts: \textbf{known mIoU / unknown mIoU / mIoU}.
    The ``Gro.'' here stands for Grounding.
    }
    \label{tab:alignment_ablation}
    \vspace{-10pt}
\end{table}

\noindent \textbf{Training paradigm}
\Cref{tab:alignment_ablation} compares different training paradigms in both closed- and open-world settings.
Similarity-based alignment with image embeddings from pre-trained \ac{VLMs} is currently the common practice for self-supervised occupancy prediction.
To analyze its effectiveness, we replace $\mathcal{L}_{\text{Grounding}}$ in AGO with a cosine similarity loss and remove the modality adapter (denoted as ``Align'').
Additionally, we evaluate a pure grounding approach without the modality adapter (``Gro.'') and a hybrid approach that applies both grounding and alignment to the same 3D embeddings (``Gro.+Align'').  
As shown in \Cref{tab:alignment_ablation}, alignment alone is not competitive in either setting, as the modality gap, low resolution, and incomplete scene representation of image embeddings limit prediction capability
Grounding alone improves known category prediction but almost fails for unknown objects, yielding unknown mIoU scores of only 0.0 and 0.3 in pretraining and zero-shot evaluation.
When both grounding and alignment are applied to the same embedding, the 3D representation is forced to adapt to text and image embeddings simultaneously.
This conflict results in a 0.19 mIoU drop in the self-supervised setting and over a 1 mIoU decrease in known categories for the open-world task.  
These findings indicate that simply combining losses is insufficient to address modality conflicts.
In contrast, our proposed adaptive grounding alleviates these gaps, leading to more robust prediction performance in both closed- and open-world scenarios.

\begin{table}
    \centering
    \footnotesize
    \resizebox{\linewidth}{!}{
        \begin{tabular}{c|c|cc}
            \toprule
            \multirow{2}{*}{\makecell{Open World \\ Identifier}} & \multirow{2}{*}{\makecell{Discrimination \\ Criteria}} & \multicolumn{2}{c}{Open World} \\
            & & Pretraining mIoU & Zero-shot mIoU\\
            \midrule
            \ding{55} & None & 22.2\,/\,1.1\,/\,11.7 & \textbf{32.8}\,/\,1.6\,/\,7.8 \\
            \midrule
            \checkmark & Max Confidence & \textbf{22.4}\,/\,3.1\,/\,12.8 & 32.6\,/\,2.8\,/\,8.7 \\
            \checkmark & Min Entropy & 22.1\,/\,\textbf{3.6}\,/\,\textbf{12.9} & 32.2\,/\,\textbf{3.2}\,/\,\textbf{9.0} \\
            \bottomrule
        \end{tabular}
    }
    \vspace{-5pt}
    \caption{
    \textbf{Ablation study of open-world inference strategy.}
    }
    \label{tab:ow_strategy_ablation}
    \vspace{-5pt}
\end{table}

\begin{table}
    \centering
    \footnotesize
    \resizebox{\linewidth}{!}{
        \begin{tabular}{c|cc|ll}
            \toprule
            \multirow{2}{*}{Method} & \multirow{2}{*}{Noise Prompts} & \multirow{2}{*}{$\mathcal{L}_{\text{Occ}}$} & \multicolumn{2}{c}{Self-supervised} \\
            & & & \multicolumn{1}{c}{IoU} & \multicolumn{1}{c}{mIoU} \\
            \midrule  
            AGO  &   &   & 53.10 & 18.01 \\ 
            \midrule  
            AGO  &   & \checkmark & 55.18 \textcolor[rgb]{0,0.5,0}{(+2.08)} & 18.97 \textcolor[rgb]{0,0.5,0}{(+0.96)}\\
            AGO  & \checkmark  & \checkmark  & \textbf{55.45} \textcolor[rgb]{0,0.5,0}{(+0.27)} & \textbf{19.23} \textcolor[rgb]{0,0.5,0}{(+0.26)}\\ 
            \bottomrule
        \end{tabular}
    }
    \vspace{-5pt}
    \caption{
    \textbf{Ablation study of architecture components.}
    }
    \label{tab:component_ablation}
    \vspace{-10pt}
\end{table}

\begin{figure}[t]
    \centering
    \includegraphics[width=\linewidth]{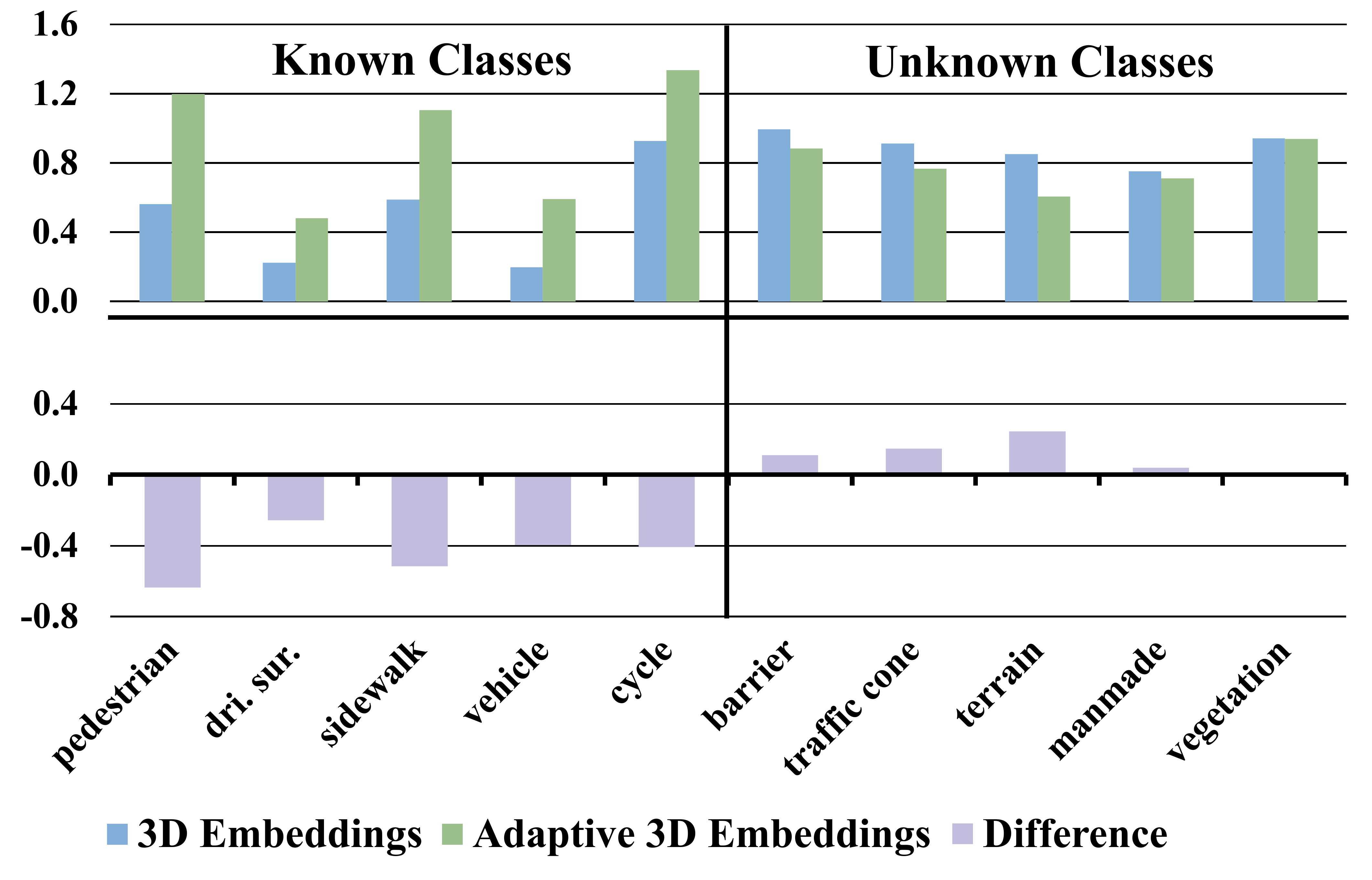}
    \vspace{-20pt}
    \caption{
    \textbf{Average information entropy for each class.}
    }
    \label{fig:ave_entropy}
    \vspace{-15pt}
\end{figure}

\noindent \textbf{Open-world inference strategy}
\Cref{tab:ow_strategy_ablation} compares the impact of the open world identifier and different criteria. 
Since fine-tuning provides pseudo-labels for all categories, we focus on pre-training and zero-shot evaluation.
Even without an identifier, the model demonstrates some open-world prediction capability through adaptive grounding.
We also observe differences in prediction distributions between the original and adaptive 3D embeddings.
As shown in~\Cref{fig:ave_entropy}, adaptive 3D embeddings generally exhibit lower information entropy for unknown categories.
Similar difference also exists in the maximum confidence distribution.
Based on this, we design an open world identifier with two criteria to further leverage the general perception capability of adaptive 3D embeddings (see the supplementary material for further details.). 
As shown in \Cref{tab:ow_strategy_ablation}, both criteria improve unknown category prediction while minimally affecting known categories.
The minimum entropy criterion provides a slightly higher gain (3.6 vs. 3.1 unknown mIoU in pretraining) compared to the confidence-based approach.

\noindent \textbf{Architecture components}
\Cref{tab:component_ablation} highlights the impact of $\mathcal{L}_{\text{Occ}}$ and the use of additional random noise prompts.
Since these components affect overall occupancy prediction performance regardless of task settings, we report comparisons based on self-supervised training.
Although post-processing techniques such as multi-frame aggregation and ray casting are applied, the pseudo-labels remain spatially incomplete.
The occupancy loss, \(\mathcal{L}_{\text{Occ}}\), helps balance the free and occupied voxels during training, leading to improvements in both IoU and mIoU.
The incorporation of noise prompts as negative samples in grounding training optimizes the semantic space, thereby enhancing the discriminative capability of the 3D embeddings.
Random sampling strategy from a general dictionary further reduces potential conflicts between noise prompts and label prompts, resulting in a 0.26 mIoU improvement.
\section{Conclusion}
\label{sec:conclusion}
In this paper, we proposed AGO, an effective 3D semantic occupancy prediction framework for open-world scenarios in autonomous driving.
By integrating grounding training with noise prompts, our AGO refines 3D features to achieve enhanced discriminative power. 
The adaptive projection effectively transfers knowledge from pretrained \ac{VLMs}, avoiding the impact of the modality gaps, while an open-world identifier based on information entropy ensures robust voxel selection for both known and unknown objects. 
Experiments on the Occ3D-nuScenes dataset demonstrate that AGO outperforms previous methods in both closed-world self-supervised tasks and open-world settings.
While promising, there are still research directions to be explored in the future, such as better integration of temporal sequence information and the usage of semantic-rich text prompts for grounding training.
Nevertheless, AGO offers a promising solution for 3D scene understanding in real-world autonomous driving applications.

\section*{Acknowledgement}
This work is a result of the joint research project STADT:up (Förderkennzeichen 19A22006O). 
The project is supported by the German Federal Ministry for Economic Affairs and Climate Action (BMWK), based on a decision of the German Bundestag. 
The author is solely responsible for the content of this publication. 
{
    \small
    \bibliographystyle{ieeenat_fullname}
    \bibliography{main}

\begin{thebibliography}{65}
\providecommand{\natexlab}[1]{#1}
\providecommand{\url}[1]{\texttt{#1}}
\expandafter\ifx\csname urlstyle\endcsname\relax
  \providecommand{\doi}[1]{doi: #1}\else
  \providecommand{\doi}{doi: \begingroup \urlstyle{rm}\Url}\fi

\bibitem[Abualhanud et~al.(2024)Abualhanud, Erahan, and Mehltretter]{abualhanud2024self}
S Abualhanud, E Erahan, and M Mehltretter.
\newblock Self-supervised 3d semantic occupancy prediction from multi-view 2d surround images.
\newblock \emph{PFG--Journal of Photogrammetry, Remote Sensing and Geoinformation Science}, 92\penalty0 (5):\penalty0 483--498, 2024.

\bibitem[Berman et~al.(2018)Berman, Triki, and Blaschko]{berman2018lovasz}
Maxim Berman, Amal~Rannen Triki, and Matthew~B Blaschko.
\newblock The lov{\'a}sz-softmax loss: A tractable surrogate for the optimization of the intersection-over-union measure in neural networks.
\newblock In \emph{Proceedings of the IEEE conference on computer vision and pattern recognition}, pages 4413--4421, 2018.

\bibitem[Boeder et~al.(2024{\natexlab{a}})Boeder, Gigengack, and Risse]{boeder2024langocc}
Simon Boeder, Fabian Gigengack, and Benjamin Risse.
\newblock Langocc: Self-supervised open vocabulary occupancy estimation via volume rendering.
\newblock \emph{arXiv preprint arXiv:2407.17310}, 2024{\natexlab{a}}.

\bibitem[Boeder et~al.(2024{\natexlab{b}})Boeder, Gigengack, and Risse]{boeder2024occflownet}
Simon Boeder, Fabian Gigengack, and Benjamin Risse.
\newblock Occflownet: Towards self-supervised occupancy estimation via differentiable rendering and occupancy flow.
\newblock \emph{arXiv preprint arXiv:2402.12792}, 2024{\natexlab{b}}.

\bibitem[Caesar et~al.(2020)Caesar, Bankiti, Lang, Vora, Liong, Xu, Krishnan, Pan, Baldan, and Beijbom]{caesar2020nuscenes}
Holger Caesar, Varun Bankiti, Alex~H Lang, Sourabh Vora, Venice~Erin Liong, Qiang Xu, Anush Krishnan, Yu Pan, Giancarlo Baldan, and Oscar Beijbom.
\newblock nuscenes: A multimodal dataset for autonomous driving.
\newblock In \emph{Proceedings of the IEEE/CVF conference on computer vision and pattern recognition}, pages 11621--11631, 2020.

\bibitem[Cao and De~Charette(2022)]{cao2022monoscene}
Anh-Quan Cao and Raoul De~Charette.
\newblock Monoscene: Monocular 3d semantic scene completion.
\newblock In \emph{Proceedings of the IEEE/CVF Conference on Computer Vision and Pattern Recognition}, pages 3991--4001, 2022.

\bibitem[Cao and de~Charette(2023)]{cao2023scenerf}
Anh-Quan Cao and Raoul de Charette.
\newblock Scenerf: Self-supervised monocular 3d scene reconstruction with radiance fields.
\newblock In \emph{Proceedings of the IEEE/CVF International Conference on Computer Vision}, pages 9387--9398, 2023.

\bibitem[Chen et~al.(2020)Chen, Li, Yu, El~Kholy, Ahmed, Gan, Cheng, and Liu]{chen2020uniter}
Yen-Chun Chen, Linjie Li, Licheng Yu, Ahmed El~Kholy, Faisal Ahmed, Zhe Gan, Yu Cheng, and Jingjing Liu.
\newblock Uniter: Universal image-text representation learning.
\newblock In \emph{European conference on computer vision}, pages 104--120. Springer, 2020.

\bibitem[Choy et~al.(2019)Choy, Gwak, and Savarese]{choy20194d}
Christopher Choy, JunYoung Gwak, and Silvio Savarese.
\newblock 4d spatio-temporal convnets: Minkowski convolutional neural networks.
\newblock In \emph{Proceedings of the IEEE/CVF conference on computer vision and pattern recognition}, pages 3075--3084, 2019.

\bibitem[Devlin et~al.(2018)Devlin, Chang, Lee, and Toutanova]{devlin2018bert}
Jacob Devlin, Ming-Wei Chang, Kenton Lee, and Kristina Toutanova.
\newblock Bert: Pre-training of deep bidirectional transformers for language understanding.
\newblock \emph{arXiv preprint arXiv:1810.04805}, 2018.

\bibitem[Ding et~al.(2025)Ding, Yang, Wiederer, Braun, Li, Gall, and Yang]{ding2025tqd}
Shuxiao Ding, Yutong Yang, Julian Wiederer, Markus Braun, Peizheng Li, Juergen Gall, and Bin Yang.
\newblock Tqd-track: Temporal query denoising for 3d multi-object tracking.
\newblock \emph{arXiv preprint arXiv:2504.03258}, 2025.

\bibitem[Fei et~al.(2020)Fei, Chen, Heidenreich, Wirges, and Stiller]{semanticvoxel}
Juncong Fei, Wenbo Chen, Philipp Heidenreich, Sascha Wirges, and Christoph Stiller.
\newblock Semanticvoxels: Sequential fusion for 3d pedestrian detection using lidar point cloud and semantic segmentation.
\newblock In \emph{2020 IEEE International Conference on Multisensor Fusion and Integration for Intelligent Systems (MFI)}, pages 185--190, 2020.

\bibitem[Fei et~al.(2021)Fei, Peng, Heidenreich, Bieder, and Stiller]{pillarseg}
Juncong Fei, Kunyu Peng, Philipp Heidenreich, Frank Bieder, and Christoph Stiller.
\newblock Pillarsegnet: Pillar-based semantic grid map estimation using sparse lidar data.
\newblock In \emph{2021 IEEE Intelligent Vehicles Symposium (IV)}, pages 838--844, 2021.

\bibitem[Fu et~al.(2024)Fu, Hamilton, Brandt, Feldman, Zhang, and Freeman]{fu2024featup}
Stephanie Fu, Mark Hamilton, Laura Brandt, Axel Feldman, Zhoutong Zhang, and William~T Freeman.
\newblock Featup: A model-agnostic framework for features at any resolution.
\newblock \emph{arXiv preprint arXiv:2403.10516}, 2024.

\bibitem[Gan et~al.(2023)Gan, Mo, Xu, and Yokoya]{gan2023simple}
Wanshui Gan, Ningkai Mo, Hongbin Xu, and Naoto Yokoya.
\newblock A simple attempt for 3d occupancy estimation in autonomous driving.
\newblock \emph{CoRR}, 2023.

\bibitem[Gan et~al.(2024)Gan, Liu, Xu, Mo, and Yokoya]{gan2024gaussianoccfullyselfsupervisedefficient}
Wanshui Gan, Fang Liu, Hongbin Xu, Ningkai Mo, and Naoto Yokoya.
\newblock Gaussianocc: Fully self-supervised and efficient 3d occupancy estimation with gaussian splatting, 2024.

\bibitem[He et~al.(2016)He, Zhang, Ren, and Sun]{he2016deep}
Kaiming He, Xiangyu Zhang, Shaoqing Ren, and Jian Sun.
\newblock Deep residual learning for image recognition.
\newblock In \emph{Proceedings of the IEEE conference on computer vision and pattern recognition}, pages 770--778, 2016.

\bibitem[Huang et~al.(2021)Huang, Huang, Zhu, Ye, and Du]{huang2021bevdet}
Junjie Huang, Guan Huang, Zheng Zhu, Yun Ye, and Dalong Du.
\newblock Bevdet: High-performance multi-camera 3d object detection in bird-eye-view.
\newblock \emph{arXiv preprint arXiv:2112.11790}, 2021.

\bibitem[Huang et~al.(2023)Huang, Zheng, Zhang, Zhou, and Lu]{huang2023tri}
Yuanhui Huang, Wenzhao Zheng, Yunpeng Zhang, Jie Zhou, and Jiwen Lu.
\newblock Tri-perspective view for vision-based 3d semantic occupancy prediction.
\newblock In \emph{Proceedings of the IEEE/CVF conference on computer vision and pattern recognition}, pages 9223--9232, 2023.

\bibitem[Huang et~al.(2024)Huang, Zheng, Zhang, Zhou, and Lu]{huang2024selfocc}
Yuanhui Huang, Wenzhao Zheng, Borui Zhang, Jie Zhou, and Jiwen Lu.
\newblock Selfocc: Self-supervised vision-based 3d occupancy prediction.
\newblock In \emph{Proceedings of the IEEE/CVF Conference on Computer Vision and Pattern Recognition}, pages 19946--19956, 2024.

\bibitem[Jiang et~al.(2024)Jiang, Liu, Cheng, Wang, Lin, Su, Liu, and Wang]{jiang2024gausstr}
Haoyi Jiang, Liu Liu, Tianheng Cheng, Xinjie Wang, Tianwei Lin, Zhizhong Su, Wenyu Liu, and Xinggang Wang.
\newblock Gausstr: Foundation model-aligned gaussian transformer for self-supervised 3d spatial understanding.
\newblock \emph{arXiv preprint arXiv:2412.13193}, 2024.

\bibitem[Jiang et~al.(2023)Jiang, Zhang, Miao, Zhu, Gao, Hu, and Jiang]{jiang2023polarformer}
Yanqin Jiang, Li Zhang, Zhenwei Miao, Xiatian Zhu, Jin Gao, Weiming Hu, and Yu-Gang Jiang.
\newblock Polarformer: Multi-camera 3d object detection with polar transformer.
\newblock In \emph{Proceedings of the AAAI conference on Artificial Intelligence}, pages 1042--1050, 2023.

\bibitem[Kirillov et~al.(2023)Kirillov, Mintun, Ravi, Mao, Rolland, Gustafson, Xiao, Whitehead, Berg, Lo, et~al.]{kirillov2023segment}
Alexander Kirillov, Eric Mintun, Nikhila Ravi, Hanzi Mao, Chloe Rolland, Laura Gustafson, Tete Xiao, Spencer Whitehead, Alexander~C Berg, Wan-Yen Lo, et~al.
\newblock Segment anything.
\newblock In \emph{Proceedings of the IEEE/CVF International Conference on Computer Vision}, pages 4015--4026, 2023.

\bibitem[Lang et~al.(2019)Lang, Vora, Caesar, Zhou, Yang, and Beijbom]{lang2019pointpillars}
Alex~H Lang, Sourabh Vora, Holger Caesar, Lubing Zhou, Jiong Yang, and Oscar Beijbom.
\newblock Pointpillars: Fast encoders for object detection from point clouds.
\newblock In \emph{Proceedings of the IEEE/CVF conference on computer vision and pattern recognition}, pages 12697--12705, 2019.

\bibitem[Li et~al.(2019)Li, Yatskar, Yin, Hsieh, and Chang]{li2019visualbert}
Liunian~Harold Li, Mark Yatskar, Da Yin, Cho-Jui Hsieh, and Kai-Wei Chang.
\newblock Visualbert: A simple and performant baseline for vision and language.
\newblock \emph{arXiv preprint arXiv:1908.03557}, 2019.

\bibitem[Li et~al.(2022{\natexlab{a}})Li, Zhang, Zhang, Yang, Li, Zhong, Wang, Yuan, Zhang, Hwang, et~al.]{li2022grounded}
Liunian~Harold Li, Pengchuan Zhang, Haotian Zhang, Jianwei Yang, Chunyuan Li, Yiwu Zhong, Lijuan Wang, Lu Yuan, Lei Zhang, Jenq-Neng Hwang, et~al.
\newblock Grounded language-image pre-training.
\newblock In \emph{Proceedings of the IEEE/CVF Conference on Computer Vision and Pattern Recognition}, pages 10965--10975, 2022{\natexlab{a}}.

\bibitem[Li et~al.(2023{\natexlab{a}})Li, Ding, Chen, Hanselmann, Cordts, and Gall]{li2023powerbev}
Peizheng Li, Shuxiao Ding, Xieyuanli Chen, Niklas Hanselmann, Marius Cordts, and Juergen Gall.
\newblock Powerbev: a powerful yet lightweight framework for instance prediction in bird's-eye view.
\newblock \emph{arXiv preprint arXiv:2306.10761}, 2023{\natexlab{a}}.

\bibitem[Li et~al.(2023{\natexlab{b}})Li, Yu, Choy, Xiao, Alvarez, Fidler, Feng, and Anandkumar]{li2023voxformer}
Yiming Li, Zhiding Yu, Christopher Choy, Chaowei Xiao, Jose~M Alvarez, Sanja Fidler, Chen Feng, and Anima Anandkumar.
\newblock Voxformer: Sparse voxel transformer for camera-based 3d semantic scene completion.
\newblock In \emph{Proceedings of the IEEE/CVF conference on computer vision and pattern recognition}, pages 9087--9098, 2023{\natexlab{b}}.

\bibitem[Li et~al.(2022{\natexlab{b}})Li, Wang, Li, Xie, Sima, Lu, Qiao, and Dai]{li2022bevformer}
Zhiqi Li, Wenhai Wang, Hongyang Li, Enze Xie, Chonghao Sima, Tong Lu, Yu Qiao, and Jifeng Dai.
\newblock Bevformer: Learning bird’s-eye-view representation from multi-camera images via spatiotemporal transformers.
\newblock In \emph{European conference on computer vision}, pages 1--18. Springer, 2022{\natexlab{b}}.

\bibitem[Lin et~al.(2024)Lin, Jin, Wang, Wei, and Dong]{lin2024teocc}
Zhiwei Lin, Hongbo Jin, Yongtao Wang, Yufei Wei, and Nan Dong.
\newblock Teocc: Radar-camera multi-modal occupancy prediction via temporal enhancement.
\newblock In \emph{ECAI 2024}, pages 129--136. IOS Press, 2024.

\bibitem[Liu et~al.(2023)Liu, Zeng, Ren, Li, Zhang, Yang, Li, Yang, Su, Zhu, et~al.]{liu2023grounding}
Shilong Liu, Zhaoyang Zeng, Tianhe Ren, Feng Li, Hao Zhang, Jie Yang, Chunyuan Li, Jianwei Yang, Hang Su, Jun Zhu, et~al.
\newblock Grounding dino: Marrying dino with grounded pre-training for open-set object detection.
\newblock \emph{arXiv preprint arXiv:2303.05499}, 2023.

\bibitem[Liu et~al.(2022)Liu, Wang, Zhang, and Sun]{liu2022petr}
Yingfei Liu, Tiancai Wang, Xiangyu Zhang, and Jian Sun.
\newblock Petr: Position embedding transformation for multi-view 3d object detection.
\newblock In \emph{European Conference on Computer Vision}, pages 531--548. Springer, 2022.

\bibitem[Liu et~al.(2024)Liu, Mou, Yu, Han, Mao, Xiong, and Wang]{liu2024let}
Yili Liu, Linzhan Mou, Xuan Yu, Chenrui Han, Sitong Mao, Rong Xiong, and Yue Wang.
\newblock Let occ flow: Self-supervised 3d occupancy flow prediction.
\newblock \emph{arXiv preprint arXiv:2407.07587}, 2024.

\bibitem[Loshchilov(2017)]{loshchilov2017decoupled}
I Loshchilov.
\newblock Decoupled weight decay regularization.
\newblock \emph{arXiv preprint arXiv:1711.05101}, 2017.

\bibitem[Lu et~al.(2019)Lu, Batra, Parikh, and Lee]{lu2019vilbert}
Jiasen Lu, Dhruv Batra, Devi Parikh, and Stefan Lee.
\newblock Vilbert: Pretraining task-agnostic visiolinguistic representations for vision-and-language tasks.
\newblock \emph{Advances in neural information processing systems}, 32, 2019.

\bibitem[Ma et~al.(2024)Ma, Chen, Huang, Xu, Luo, Xu, Gu, Ai, and Wang]{ma2024cam4docc}
Junyi Ma, Xieyuanli Chen, Jiawei Huang, Jingyi Xu, Zhen Luo, Jintao Xu, Weihao Gu, Rui Ai, and Hesheng Wang.
\newblock Cam4docc: Benchmark for camera-only 4d occupancy forecasting in autonomous driving applications.
\newblock In \emph{Proceedings of the IEEE/CVF Conference on Computer Vision and Pattern Recognition}, pages 21486--21495, 2024.

\bibitem[Philion and Fidler(2020)]{philion2020lift}
Jonah Philion and Sanja Fidler.
\newblock Lift, splat, shoot: Encoding images from arbitrary camera rigs by implicitly unprojecting to 3d.
\newblock In \emph{Computer Vision--ECCV 2020: 16th European Conference, Glasgow, UK, August 23--28, 2020, Proceedings, Part XIV 16}, pages 194--210. Springer, 2020.

\bibitem[Radford et~al.(2021)Radford, Kim, Hallacy, Ramesh, Goh, Agarwal, Sastry, Askell, Mishkin, Clark, et~al.]{radford2021learning}
Alec Radford, Jong~Wook Kim, Chris Hallacy, Aditya Ramesh, Gabriel Goh, Sandhini Agarwal, Girish Sastry, Amanda Askell, Pamela Mishkin, Jack Clark, et~al.
\newblock Learning transferable visual models from natural language supervision.
\newblock In \emph{International conference on machine learning}, pages 8748--8763. PMLR, 2021.

\bibitem[Ranftl et~al.(2020)Ranftl, Lasinger, Hafner, Schindler, and Koltun]{ranftl2020towards}
Ren{\'e} Ranftl, Katrin Lasinger, David Hafner, Konrad Schindler, and Vladlen Koltun.
\newblock Towards robust monocular depth estimation: Mixing datasets for zero-shot cross-dataset transfer.
\newblock \emph{IEEE transactions on pattern analysis and machine intelligence}, 44\penalty0 (3):\penalty0 1623--1637, 2020.

\bibitem[Ren et~al.(2024)Ren, Liu, Zeng, Lin, Li, Cao, Chen, Huang, Chen, Yan, et~al.]{ren2024grounded}
Tianhe Ren, Shilong Liu, Ailing Zeng, Jing Lin, Kunchang Li, He Cao, Jiayu Chen, Xinyu Huang, Yukang Chen, Feng Yan, et~al.
\newblock Grounded sam: Assembling open-world models for diverse visual tasks.
\newblock \emph{arXiv preprint arXiv:2401.14159}, 2024.

\bibitem[Roldao et~al.(2020)Roldao, de~Charette, and Verroust-Blondet]{roldao2020lmscnet}
Luis Roldao, Raoul de Charette, and Anne Verroust-Blondet.
\newblock Lmscnet: Lightweight multiscale 3d semantic completion.
\newblock In \emph{2020 International Conference on 3D Vision (3DV)}, pages 111--119. IEEE, 2020.

\bibitem[Shi et~al.(2024)Shi, Cheng, Zhang, Liu, and Wang]{shi2024occupancy}
Yiang Shi, Tianheng Cheng, Qian Zhang, Wenyu Liu, and Xinggang Wang.
\newblock Occupancy as set of points.
\newblock In \emph{European Conference on Computer Vision}, pages 72--87. Springer, 2024.

\bibitem[Tan and Bansal(2019)]{tan2019lxmert}
Hao Tan and Mohit Bansal.
\newblock Lxmert: Learning cross-modality encoder representations from transformers.
\newblock \emph{arXiv preprint arXiv:1908.07490}, 2019.

\bibitem[Tan et~al.(2023)Tan, Dong, Zhang, Zhang, Ji, and Li]{tan2023ovo}
Zhiyu Tan, Zichao Dong, Cheng Zhang, Weikun Zhang, Hang Ji, and Hao Li.
\newblock Ovo: Open-vocabulary occupancy.
\newblock \emph{arXiv preprint arXiv:2305.16133}, 2023.

\bibitem[Tian et~al.(2024)Tian, Jiang, Yun, Mao, Yang, Wang, Wang, and Zhao]{tian2024occ3d}
Xiaoyu Tian, Tao Jiang, Longfei Yun, Yucheng Mao, Huitong Yang, Yue Wang, Yilun Wang, and Hang Zhao.
\newblock Occ3d: A large-scale 3d occupancy prediction benchmark for autonomous driving.
\newblock \emph{Advances in Neural Information Processing Systems}, 36, 2024.

\bibitem[Tong et~al.(2023)Tong, Sima, Wang, Chen, Wu, Deng, Gu, Lu, Luo, Lin, et~al.]{tong2023scene}
Wenwen Tong, Chonghao Sima, Tai Wang, Li Chen, Silei Wu, Hanming Deng, Yi Gu, Lewei Lu, Ping Luo, Dahua Lin, et~al.
\newblock Scene as occupancy.
\newblock In \emph{Proceedings of the IEEE/CVF International Conference on Computer Vision}, pages 8406--8415, 2023.

\bibitem[Vobecky et~al.(2024)Vobecky, Sim{\'e}oni, Hurych, Gidaris, Bursuc, P{\'e}rez, and Sivic]{vobecky2024pop}
Antonin Vobecky, Oriane Sim{\'e}oni, David Hurych, Spyridon Gidaris, Andrei Bursuc, Patrick P{\'e}rez, and Josef Sivic.
\newblock Pop-3d: Open-vocabulary 3d occupancy prediction from images.
\newblock \emph{Advances in Neural Information Processing Systems}, 36, 2024.

\bibitem[Wang et~al.(2022)Wang, Guizilini, Zhang, Wang, Zhao, and Solomon]{wang2022detr3d}
Yue Wang, Vitor~Campagnolo Guizilini, Tianyuan Zhang, Yilun Wang, Hang Zhao, and Justin Solomon.
\newblock Detr3d: 3d object detection from multi-view images via 3d-to-2d queries.
\newblock In \emph{Conference on Robot Learning}, pages 180--191. PMLR, 2022.

\bibitem[Wang et~al.(2024)Wang, Chen, Liao, Fan, and Zhang]{wang2024panoocc}
Yuqi Wang, Yuntao Chen, Xingyu Liao, Lue Fan, and Zhaoxiang Zhang.
\newblock Panoocc: Unified occupancy representation for camera-based 3d panoptic segmentation.
\newblock In \emph{Proceedings of the IEEE/CVF conference on computer vision and pattern recognition}, pages 17158--17168, 2024.

\bibitem[Wang et~al.(2025)Wang, Huang, Sun, Yan, Xing, Tu, and Li]{wang2025uniocc}
Yuping Wang, Xiangyu Huang, Xiaokang Sun, Mingxuan Yan, Shuo Xing, Zhengzhong Tu, and Jiachen Li.
\newblock Uniocc: A unified benchmark for occupancy forecasting and prediction in autonomous driving.
\newblock \emph{arXiv preprint arXiv:2503.24381}, 2025.

\bibitem[Wei et~al.(2023)Wei, Zhao, Zheng, Zhu, Zhou, and Lu]{wei2023surroundocc}
Yi Wei, Linqing Zhao, Wenzhao Zheng, Zheng Zhu, Jie Zhou, and Jiwen Lu.
\newblock Surroundocc: Multi-camera 3d occupancy prediction for autonomous driving.
\newblock In \emph{Proceedings of the IEEE/CVF International Conference on Computer Vision}, pages 21729--21740, 2023.

\bibitem[Xia et~al.(2023)Xia, Liu, Li, Zhu, Ma, Li, Hou, and Qiao]{xia2023scpnet}
Zhaoyang Xia, Youquan Liu, Xin Li, Xinge Zhu, Yuexin Ma, Yikang Li, Yuenan Hou, and Yu Qiao.
\newblock Scpnet: Semantic scene completion on point cloud.
\newblock In \emph{Proceedings of the IEEE/CVF conference on computer vision and pattern recognition}, pages 17642--17651, 2023.

\bibitem[Yan et~al.(2024)Yan, Dong, Shao, Lu, Haiyang, Liu, Wang, Wang, Wang, Remondino, et~al.]{yan2024renderworld}
Ziyang Yan, Wenzhen Dong, Yihua Shao, Yuhang Lu, Liu Haiyang, Jingwen Liu, Haozhe Wang, Zhe Wang, Yan Wang, Fabio Remondino, et~al.
\newblock Renderworld: World model with self-supervised 3d label.
\newblock \emph{arXiv preprint arXiv:2409.11356}, 2024.

\bibitem[Yin et~al.(2021)Yin, Zhou, and Krahenbuhl]{yin2021center}
Tianwei Yin, Xingyi Zhou, and Philipp Krahenbuhl.
\newblock Center-based 3d object detection and tracking.
\newblock In \emph{Proceedings of the IEEE/CVF conference on computer vision and pattern recognition}, pages 11784--11793, 2021.

\bibitem[Zhang et~al.(2023{\natexlab{a}})Zhang, Yan, Wei, Li, Liu, Tang, Duan, and Lu]{zhang2023occnerf}
Chubin Zhang, Juncheng Yan, Yi Wei, Jiaxin Li, Li Liu, Yansong Tang, Yueqi Duan, and Jiwen Lu.
\newblock Occnerf: Self-supervised multi-camera occupancy prediction with neural radiance fields.
\newblock \emph{arXiv e-prints}, pages arXiv--2312, 2023{\natexlab{a}}.

\bibitem[Zhang et~al.(2024{\natexlab{a}})Zhang, Huang, Jin, and Lu]{zhang2024vision}
Jingyi Zhang, Jiaxing Huang, Sheng Jin, and Shijian Lu.
\newblock Vision-language models for vision tasks: A survey.
\newblock \emph{IEEE Transactions on Pattern Analysis and Machine Intelligence}, 2024{\natexlab{a}}.

\bibitem[Zhang et~al.(2024{\natexlab{b}})Zhang, Yang, Fang, Geng, and Jensfelt]{zhang2024deflow}
Qingwen Zhang, Yi Yang, Heng Fang, Ruoyu Geng, and Patric Jensfelt.
\newblock Deflow: Decoder of scene flow network in autonomous driving.
\newblock \emph{arXiv preprint arXiv:2401.16122}, 2024{\natexlab{b}}.

\bibitem[Zhang et~al.(2025)Zhang, Yang, Li, Andersson, and Jensfelt]{zhang2025seflow}
Qingwen Zhang, Yi Yang, Peizheng Li, Olov Andersson, and Patric Jensfelt.
\newblock Seflow: A self-supervised scene flow method in autonomous driving.
\newblock In \emph{European Conference on Computer Vision}, pages 353--369. Springer, 2025.

\bibitem[Zhang et~al.(2023{\natexlab{b}})Zhang, Zhu, and Du]{zhang2023occformer}
Yunpeng Zhang, Zheng Zhu, and Dalong Du.
\newblock Occformer: Dual-path transformer for vision-based 3d semantic occupancy prediction.
\newblock In \emph{Proceedings of the IEEE/CVF International Conference on Computer Vision}, pages 9433--9443, 2023{\natexlab{b}}.

\bibitem[Zhao et~al.(2024)Zhao, Chen, Sun, Yang, Wang, Zhang, Li, Kou, Wei, and Zhang]{zhao2024hybridocc}
Xiao Zhao, Bo Chen, Mingyang Sun, Dingkang Yang, Youxing Wang, Xukun Zhang, Mingcheng Li, Dongliang Kou, Xiaoyi Wei, and Lihua Zhang.
\newblock Hybridocc: Nerf enhanced transformer-based multi-camera 3d occupancy prediction.
\newblock \emph{IEEE Robotics and Automation Letters}, 2024.

\bibitem[Zheng et~al.(2024)Zheng, Tang, Wang, Wang, Ren, Feng, and Ma]{zheng2024veon}
Jilai Zheng, Pin Tang, Zhongdao Wang, Guoqing Wang, Xiangxuan Ren, Bailan Feng, and Chao Ma.
\newblock Veon: Vocabulary-enhanced occupancy prediction.
\newblock \emph{arXiv preprint arXiv:2407.12294}, 2024.

\bibitem[Zhou and Kr{\"a}henb{\"u}hl(2022)]{zhou2022cross}
Brady Zhou and Philipp Kr{\"a}henb{\"u}hl.
\newblock Cross-view transformers for real-time map-view semantic segmentation.
\newblock In \emph{Proceedings of the IEEE/CVF conference on computer vision and pattern recognition}, pages 13760--13769, 2022.

\bibitem[Zhou et~al.(2022)Zhou, Loy, and Dai]{zhou2022extract}
Chong Zhou, Chen~Change Loy, and Bo Dai.
\newblock Extract free dense labels from clip.
\newblock In \emph{European Conference on Computer Vision}, pages 696--712. Springer, 2022.

\bibitem[Zhou et~al.(2025)Zhou, Wang, Wang, Wei, Dong, and Yang]{zhou2025autoocc}
Xiaoyu Zhou, Jingqi Wang, Yongtao Wang, Yufei Wei, Nan Dong, and Ming-Hsuan Yang.
\newblock Autoocc: Automatic open-ended semantic occupancy annotation via vision-language guided gaussian splatting.
\newblock \emph{arXiv preprint arXiv:2502.04981}, 2025.

\bibitem[Zhou and Tuzel(2018)]{zhou2018voxelnet}
Yin Zhou and Oncel Tuzel.
\newblock Voxelnet: End-to-end learning for point cloud based 3d object detection.
\newblock In \emph{Proceedings of the IEEE conference on computer vision and pattern recognition}, pages 4490--4499, 2018.

\end{thebibliography}
}

\clearpage
\setcounter{page}{1}
\maketitlesupplementary
\appendix
\setcounter{section}{0}
\setcounter{table}{0}
\setcounter{figure}{0}
\setcounter{equation}{0}

\newcommand{\appendixlabel}[1]{\Alph{#1}}

\renewcommand{\thesection}{\appendixlabel{section}}
\renewcommand{\thetable}{\appendixlabel{table}}
\renewcommand{\thefigure}{\appendixlabel{figure}}
\renewcommand{\theequation}{\appendixlabel{equation}}

In this supplementary material, we first describe the implementation details of our method in~\secref{sec:addl_imple_details}.
Following that, additional experimental results and ablation study are presented in~\secref{sec:addl_exp_ana}.
Finally,~\secref{sec:addl_vis} provides more visualization comparisons and qualitative analysis.

\section{Additional Implementation Details}
\label{sec:addl_imple_details}

\subsection{Label space design}
\label{subsec:label_space_design}
Choosing appropriate label prompts is crucial for effectively establishing semantic associations between categories in the language priors.
However, some of the category names in the Occ3D-nuScenes dataset~\cite{tian2024occ3d} are too vague or broad to be directly encoded into semantically rich text embeddings.
For example, in the definition of Occ3D-nuScenes~\cite{tian2024occ3d}, ``others'' and ``other flat'' represent structures and horizontal ground-level planes that cannot be classified into any other category, respectively.
They are just general terms for many categories that have not been specifically annotated and thus do not have a clear semantic meaning.
Therefore, we follow existing works~\cite{vobecky2024pop, zheng2024veon, boeder2024langocc} and divide them into the subcategories as shown in~\Cref{tab:subclass_desc}.
The above design is employed in all supervised and self-supervised experiments mentioned in this paper.
For pretraining as well as zero-shot and few-shot transfer in the open-world setting, we adopt the original labels as corresponding prompts after removing the semantically ambiguous ``others'' and ``other flat''.
The reason for this is to avoid the influence of different label space designs on open-world performance.

\begin{table*}
    \vspace{-10pt}
    \small
    \centering
    \resizebox{\textwidth}{!}{
        \begin{tabular}{l|l}
        \toprule
        Classes              & Subclasses \\
        \midrule 
        others               & \makecell[l]{animal, skateboard, segway, scooter, stroller, wheelchair, trash bag, dolley, wheel barrow, \\trash bin, shopping cart, bicycle rack, ambulance, police vehicle, cyclist}  \\
        barrier              & barrier \\
        bicycle              & bicycle \\
        bus                  & bendy bus, rigid bus \\
        car                  & car, van, suv \\
        construction vehicle & construction vehicle \\
        motorcycle           & motorcycle \\
        pedestrian           & adult pedestrian, child pedestrian, worker, police officer \\
        traffic cone         & traffic cone \\
        trailer              & trailer \\
        truck                & truck \\
        driveable surface    & road \\
        other flat           & traffic island, traffic delimiter, rail track, lake, river \\
        sidewalk             & sidewalk \\
        terrain              & lawn \\
        manmade              & building, sign, pole, traffic light \\
        vegetation           & tree, bush \\
        \bottomrule
    \end{tabular}}
    \caption{
    \textbf{Subclass description in label space design.}
    These subclasses apply to self-supervised and open-world training.
    }
    \label{tab:subclass_desc}
\end{table*}

\subsection{Pseudo-label generation}
\label{subsec:pseudo_label_generation}
Thanks to the integrated SAM~\cite{kirillov2023segment}, Grounded SAM~\cite{ren2024grounded} can generate semantic masks with more detail and more accurate boundaries than MaskCLIP+~\cite{zhou2022extract}.
Therefore, we utilize pre-trained Grounded SAM to generate pseudo-labels.
We feed each surrounding image and the label prompts defined in~\Cref{subsec:label_space_design} into this model to generate 2D pseudo semantic masks corresponding to the image.
In this process, the box threshold is set to 0.2 and the text threshold is set to 0.15.
These 2D masks are projected into the 3D voxels based on the calibration matrices of the LiDAR and surrounding cameras.

Considering the sparsity of the LiDAR point cloud, we aggregate multiple frames to densify it.
Specifically, we first select a certain number $N_\text{sweep}$ of camera sweeps and pseudo-synchronize each of them with the temporally closest LiDAR sweep.
It is worth noting that the above selection is not continuous, but has a sampling interval of $N_\text{interval}$.
This is to allow the pseudo labels to cover as much space as possible while using the same number of sweeps.
Thereafter, the generated voxelized 3D pseudo-labels of each sweep are warped to the reference sweep and superimposed to obtain dense pseudo-labels.
In our implementation, we set $N_\text{sweep}=30$ and $N_\text{interval}=2$ corresponding to a 15\,s time range and take the key frame as the reference sweep.
Despite this, there are still many false negatives in the above pseudo-labels, \ie, many occupied voxels are marked as free, which is caused by occlusion.
Therefore, we employ ray casting for free voxel assignment, that is, only unoccupied voxels between the LiDAR and the reflection point on each ray are set as free, while the remaining occluded voxels are ignored.
In addition, several LiDAR points may be located in the same voxel.
For this reason, we apply a semantic voting mechanism for each voxel, which selects the category with the most corresponding points as the category of the voxel.

\begin{figure}[t!]
    \centering
    \includegraphics[width=\linewidth]{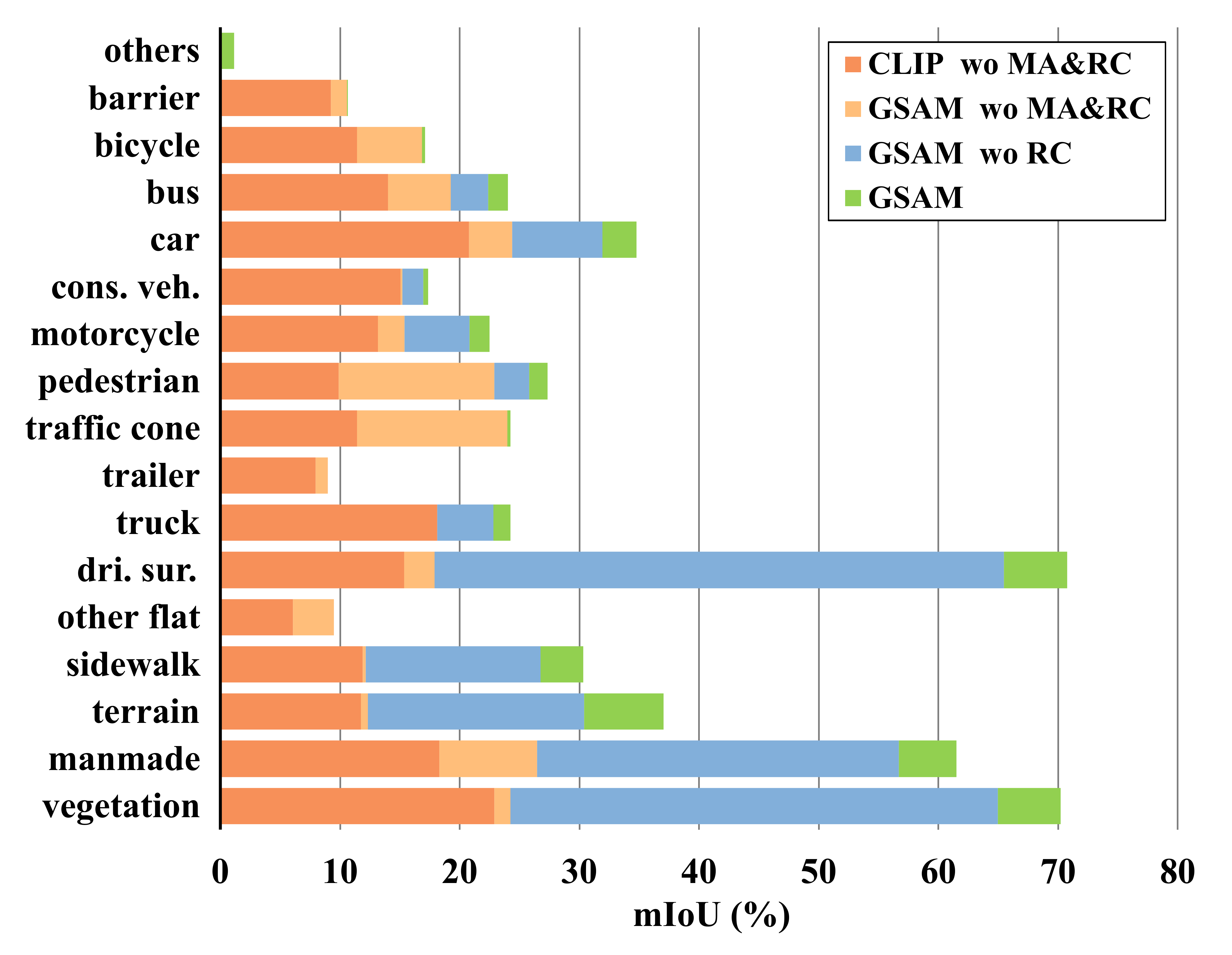}
    \vspace{-15pt}
    \caption{
    \textbf{Pseudo-label evaluation on the Occ3D-nuScenes \cite{tian2024occ3d} dataset.} 
    ``MA'' denotes multi-frame aggregation and ``RC'' refers to ray casting. 
    ``GSAM'' is an acronym for Grounded SAM.
    }
    \label{fig:pseudo_label_eval}
    \vspace{-15pt}
\end{figure}

\Cref{tab:pseudo_label_eval} compares the evaluation results of the pseudo-labels with different model bases and post-processing.
As can be seen, the mIoU of the pseudo-labels generated based on Grounded SAM is 3.24 higher than that of MaskCLIP+.
Multi-frame aggregation brings an improvement of 9.89 mIoU, while ray casting further increases the mIoU by 2.08.
It is worth noting that we did not conduct extensive prompt engineering to enhance the quality of pseudo-labels.
However, our method even outperforms the pseudo labels in many categories, such as ``driveable surface'', ``sidewalk'' and ``terrain''.
This further demonstrates the effectiveness of our proposed AGO.

\begin{table*}[htbp]
    \centering
    \resizebox{\textwidth}{!}{
    \begin{tabular}{l|c|ccccccccccccccccc|c}
        \toprule
        Method
        & Model Base
        & \rotatebox{90}{\parbox{2cm}{\raggedright\tikz\draw[fill={rgb,255: red,0; green,0; blue,0}, draw={rgb,255: red,0; green,0; blue,0}] (0,0) rectangle (0.2,0.2);~others}} 
        & \rotatebox{90}{\parbox{2cm}{\raggedright\tikz\draw[fill={rgb,255: red,255; green,120; blue,50}, draw={rgb,255: red,255; green,120; blue,50}] (0,0) rectangle (0.2,0.2);~barrier}} 
        & \rotatebox{90}{\parbox{2cm}{\raggedright\tikz\draw[fill={rgb,255: red,255; green,192; blue,203}, draw={rgb,255: red,255; green,192; blue,203}] (0,0) rectangle (0.2,0.2);~bicycle}} 
        & \rotatebox{90}{\parbox{2cm}{\raggedright\tikz\draw[fill={rgb,255: red,255; green,255; blue,0}, draw={rgb,255: red,255; green,255; blue,0}] (0,0) rectangle (0.2,0.2);~bus}}
        & \rotatebox{90}{\parbox{2cm}{\raggedright\tikz\draw[fill={rgb,255: red,0; green,150; blue,245}, draw={rgb,255: red,0; green,150; blue,245}] (0,0) rectangle (0.2,0.2);~car}} 
        & \rotatebox{90}{\raggedright\tikz\draw[fill={rgb,255: red,0; green,255; blue,255}, draw={rgb,255: red,0; green,255; blue,255}] (0,0) rectangle (0.2,0.2);~\parbox{2cm}{construction\\vehicle}}
        & \rotatebox{90}{\parbox{2cm}{\raggedright\tikz\draw[fill={rgb,255: red,200; green,180; blue,0}, draw={rgb,255: red,200; green,180; blue,0}] (0,0) rectangle (0.2,0.2);~motorcycle}} 
        & \rotatebox{90}{\parbox{2cm}{\raggedright\tikz\draw[fill={rgb,255: red,255; green,0; blue,0}, draw={rgb,255: red,255; green,0; blue,0}] (0,0) rectangle (0.2,0.2);~pedestrian}}
        & \rotatebox{90}{\parbox{2cm}{\raggedright\tikz\draw[fill={rgb,255: red,255; green,240; blue,150}, draw={rgb,255: red,255; green,240; blue,150}] (0,0) rectangle (0.2,0.2);~traffic cone}}
        & \rotatebox{90}{\parbox{2cm}{\raggedright\tikz\draw[fill={rgb,255: red,135; green,60; blue,0}, draw={rgb,255: red,135; green,60; blue,0}] (0,0) rectangle (0.2,0.2);~trailer}} 
        & \rotatebox{90}{\parbox{2cm}{\raggedright\tikz\draw[fill={rgb,255: red,160; green,32; blue,240}, draw={rgb,255: red,160; green,32; blue,240}] (0,0) rectangle (0.2,0.2);~truck}}
        & \rotatebox{90}{\raggedright\tikz\draw[fill={rgb,255: red,255; green,0; blue,255}, draw={rgb,255: red,255; green,0; blue,255}] (0,0) rectangle (0.2,0.2);~\parbox{2cm}{driveable\\surface}}
        & \rotatebox{90}{\parbox{2cm}{\raggedright\tikz\draw[fill={rgb,255: red,139; green,137; blue,137}, draw={rgb,255: red,139; green,137; blue,137}] (0,0) rectangle (0.2,0.2);~other flat}}
        & \rotatebox{90}{\parbox{2cm}{\raggedright\tikz\draw[fill={rgb,255: red,75; green,0; blue,75}, draw={rgb,255: red,75; green,0; blue,75}] (0,0) rectangle (0.2,0.2);~sidewalk}}
        & \rotatebox{90}{\parbox{2cm}{\raggedright\tikz\draw[fill={rgb,255: red,150; green,240; blue,80}, draw={rgb,255: red,150; green,240; blue,80}] (0,0) rectangle (0.2,0.2);~terrain}}
        & \rotatebox{90}{\parbox{2cm}{\raggedright\tikz\draw[fill={rgb,255: red,230; green,230; blue,250}, draw={rgb,255: red,230; green,230; blue,250}] (0,0) rectangle (0.2,0.2);~manmade}}
        & \rotatebox{90}{\parbox{2cm}{\raggedright\tikz\draw[fill={rgb,255: red,0; green,175; blue,0}, draw={rgb,255: red,0; green,175; blue,0}] (0,0) rectangle (0.2,0.2);~vegetation}}
        & \rotatebox{90}{mIoU} \\
        \midrule           
        CLIP wo MA\&RC & MaskCLIP+ & 0.05   & 9.23    & 11.41   & 13.99 & 20.79 & 15.05      & 13.15      & 9.90       & 11.43        & 7.97    & 18.11 & 15.36     & 6.08       & 11.88    & 11.74   & 18.32   & 22.90      & 12.79   \\
        GSAM wo MA\&RC & Grounded SAM & 0.04   & 10.58   & 16.86   & 19.26 & 24.37 & 15.19      & 15.38      & 22.91      & 23.98        & 8.96    & 12.37 & 17.90     & 9.48       & 12.14    & 12.34   & 26.48   & 24.24      & 16.03   \\
        GSAM wo RC & Grounded SAM & 0.03   & 10.44   & 15.09   & 22.38 & 31.94 & 16.97      & 20.81      & 25.79      & 23.55        & 7.44    & 17.09 & 65.49     & 4.84       & 26.76    & 30.39   & 56.69   & 64.95      & 25.92  \\
        GSAM & Grounded SAM & 1.12   & 10.50   & 15.33   & 24.02 & 34.77 & 17.34      & 22.50      & 27.34      & 23.82        & 6.78    & 18.51 & 70.77     & 4.14       & 30.31    & 37.03   & 61.50   & 70.21      & 28.00  \\
        \midrule   
        Self-supervised AGO & - & 1.53 & 6.75 & 6.43 & 14.00 & 22.82 & 5.57 & 16.66 & 13.20 & 6.80 & 10.53 & 15.89 & 71.48 & 4.48 & 34.48 & 41.37 & 29.33 & 25.66 & 19.23 \\
        \bottomrule
    \end{tabular}}
    \caption{
    \textbf{Pseudo-label evaluation on the Occ3D-nuScenes~\cite{tian2024occ3d} dataset.} 
    ``MA'' denotes multi-frame aggregation and ``RC'' refers to ray casting.
    ``GSAM'' is an acronym for Grounded SAM.
    }
    \label{tab:pseudo_label_eval}
\end{table*}

\subsection{Framework details}
\label{subsec:framework_details}
As described in~\Cref{subsec:model_structure}, AGO features a dual-stream architecture with a text encoder and a vision-centric 3D encoder.
Since the traditional TPVFormer~\cite{huang2023tri} is designed for the LiDAR point cloud segmentation task, it involves both point-level and voxel-level supervision.
But for 3D semantic occupancy prediction, point-level supervision is not necessary. 
Therefore, we remove this part from our implementation.

Since we split the original categories into subcategories according to~\Cref{subsec:label_space_design}, during supervised grounded training, one class in the ground truth may correspond to the text embeddings and similarity scores of multiple subclasses.
To solve this problem, for each voxel, we take the maximum score across all subclasses belonging to the same class as its similarity score.
In other words, as long as one subcategory exhibits an extremely high similarity, the original category containing that subcategory should be considered the occupancy prediction for the corresponding voxel.

In AGO, we use a dictionary obtained from Natural Language Toolkit (NLTK) library of Python as the source of noise prompts.
For each step, we randomly select $N_\text{noise}$ prompts from it and encode them into corresponding noise embeddings using the same text encoder.
In our implementation, we set $N_\text{noise}=100$.

In addition, the main purpose of the open world identifier is to flexibly select suitable features based on the prediction distribution of the original 3D embeddings and adaptive 3D embeddings.
Considering that in the closed-world setting, all categories are known during grounding training, the original 3D embedding has stronger discriminative ability.
Therefore, during closed-world prediction, the open world identifier directly selects the original 3D embedding for the final prediction.

\begin{table*}[t]
    \vspace{-5pt}
    \centering
    \footnotesize
    \begin{tabular}{c|c|c|c|c|c}
    \toprule
    Method      & Image Backbone  & Training Epochs  & Image Resolution & Self-supervised IoU & Self-supervised mIoU  \\
    \midrule 
    AGO & ResNet-101 & 24 & 900×1600 & 55.45 & 19.32 \\
    AGO & ResNet-50 & 24 & 900×1600 & 50.76 & 15.23 \\
    AGO & ResNet-50 & 12 & 900×1600 & 50.06 & 14.84 \\
    AGO & ResNet-50 & 12 & 450×800 & 50.24 & 14.78 \\
    \bottomrule
    \end{tabular}
    \vspace{-5pt}
    \caption{
    \textbf{Ablation study of image backbones, traninig epochs and resolutions.}
    }
    \label{tab:image_bb_ep_reso_ablation}
    \vspace{-10pt}
\end{table*}

\begin{table}[t]
    \centering
    \footnotesize
    \resizebox{\linewidth}{!}{
        \begin{tabular}{c|ccc}
            \toprule
            \multirow{2}{*}{\makecell{$\mathcal{L}_{\text{Alignment}}$}} & \multicolumn{3}{c}{Open World} \\
            & Pretraining mIoU & Zero-shot mIoU & Few-shot mIoU\\
            \midrule
            Cosine & 22.1\,/\,3.6\,/\,12.9 & 32.2\,/\,3.2\,/\,9.0 & 38.2\,/\,8.5\,/\,14.4\\
            \midrule
            MSE (L2) & 21.1\,/\,0.3\,/\,11.1 & 31.7\,/\,0.4\,/\,6.7 & 36.9\,/\,7.1\,/\,12.4\\
            MAE (L1) & 20.8\,/\,0.1\,/\,10.5 & 30.1\,/\,0.3\,/\,6.4 & 37.1\,/\,6.2\,/\,12.0\\
            \bottomrule
        \end{tabular}
    }
    \vspace{-5pt}
    \caption{
    \textbf{Ablation study of open-world inference strategy.}
    }
    \label{tab:align_loss_ablation}
    \vspace{-5pt}
\end{table}

\begin{table}[t]
    \centering
    \footnotesize
    \resizebox{\linewidth}{!}{
        \begin{tabular}{c|ccc}
            \toprule
            \multirow{2}{*}{\makecell{Number of \\ MLP Layers}} & \multicolumn{3}{c}{Open World} \\
            & Pretraining mIoU & Zero-shot mIoU & Few-shot mIoU\\
            \midrule
            1 & 18.4\,/\,2.6\,/\,10.5 & 29.2\,/\,1.7\,/\,6.8 & 38.0\,/\,6.5\,/\,13.2\\
            2 & 22.1\,/\,3.6\,/\,12.9 & 32.2\,/\,3.2\,/\,9.0 & 38.2\,/\,8.5\,/\,14.4\\
            3 & 14.3\,/\,3.7\,/\,10.4 & 22.0\,/\,3.4\,/\,8.3 & 37.3\,/\,8.2\,/\,13.9\\
            4 & 13.6\,/\,3.8\,/\,10.3 & 19.5\,/\,3.4\,/\,8.1 & 37.1\,/\,7.9\,/\,13.7\\
            \bottomrule
        \end{tabular}
    }
    \vspace{-5pt}
    \caption{
    \textbf{Ablation study of MLP layer number.}
    }
    \label{tab:mlp_layer_ablation}
    \vspace{-5pt}
\end{table}

\subsection{Closed-world task details}
\label{subsec:closed_world_task_details}
For the vast majority of closed-world methods, we directly adopt the performance values reported in their papers. However, POP-3D~\cite{vobecky2024pop}, as a pioneering alignment-based zero-shot method, has not yet been compared on the Occ3D~\cite{tian2024occ3d} benchmark.
Its evaluation is based on the original nuScenes dataset~\cite{caesar2020nuscenes} and considers only the voxels traversed by LiDAR rays in a single frame, with a resolution of 100×100×8 (see Sec. 4.1 in POP-3D~\cite{vobecky2024pop}).
For fairness, we retrain it using the 200×200×16 resolution consistent with Occ3D~\cite{tian2024occ3d} setting.
The drop in reported mIoU largely stems from the stricter evaluation protocol (see Sec. 3.3 \& 6.1 in Occ3D~\cite{tian2024occ3d}).

\subsection{Open-world task details}
\label{subsec:open_world_task_details}
In the open-world task, the entire progress is divided into three stages: pre-training, zero-shot evaluation and few-shot finetuning. 
In the pretraining stage, only the pseudo-labels of 5 major categories are known, namely $\mathbb{C}_{pre.} =$ \{ ``pedestrian'', ``driveable surface'', ``sidewalk'', ``vehicle'', ``cycle'' \}.
Among them, ``vehicle'' and ``cycle'' are two supercategories, which are formed by the original category sets \{ ``car'', ``bus'', ``construction vehicle'', ``trailer'', ``truck'' \} and \{ ``bicycle'', ``motorcycle'' \}, respectively, to simulate the common coarse-to-fine labeling process in real-world applications.
The remaining Occ3D classes as well as corresponding voxels are ignored during pre-training but included in the subsequent zero-shot evaluation.
During the few-shot fine-tuning stage, only a small number of samples are provided.
They have the complete original Occ3D label space and each category appears at least in $k$ samples ($k$-shot setting).
We set $k = 100$ and repeat every few-shot experiment 5 times to calculate the average, thus reducing randomness.
This is to validate model's few-shot generalization ability of unknown classes.
Noting that the original classes ``others'' and ``other flat'' are semantically ambiguous, they are not included in all stages.

Due to different categories in open-world pretraining stage, all methods need to be retrained (based on their original code).
The specific retraining settings are as follows:
\begin{itemize}
    \item POP-3D~\cite{vobecky2024pop}: It is trained solely with the original alignment loss, without pseudo-labels.
    The only difference across stages lies in the different class text prompts used for inference and evaluation.
    \item SelfOcc~\cite{huang2024selfocc}: We define its output label space as the full class set ($\mathbb{C}_k \cup \mathbb{C}_{uk}$).
    In the open-world pretraining stage, only outputs of $\mathbb{C}_k$ are supervised by pseudo-labels, while all categories are considered during zero-shot evaluation, resulting in 0 unknown mIoU.
    In the open-world fine-tuning stage, all class outputs are supervised.
    \item GaussTR~\cite{jiang2024gausstr}: The alignment loss is the same at all open-world stages.
    In the open-world pretraining stage, only pseudo-labels of $\mathbb{C}_k$ are used for the extra loss to refine the semantic boundaries, while in the open-world fine-tuning stage, pseudo-labels of all classes are used.
\end{itemize}

\section{Additional Experiments and Analyses}
\label{sec:addl_exp_ana}

\subsection{Additional ablation study}
\label{subsec:addl_ablation}

\begin{figure}[t]
    \centering
    \includegraphics[width=\linewidth]{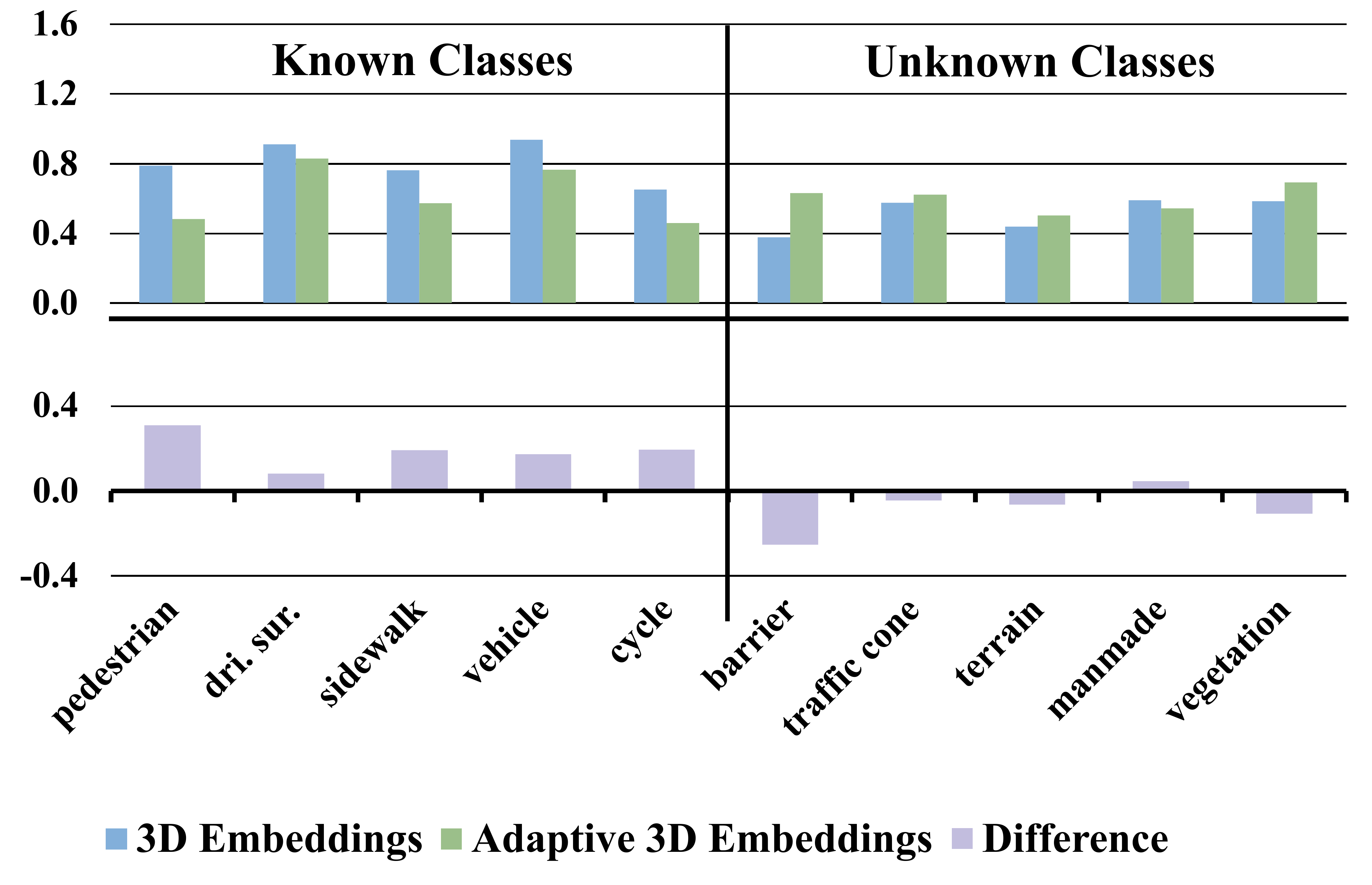}
    \caption{
    \textbf{Average maximum confidence for each class.}
    }
    \label{fig:max_confi}
    \vspace{-5pt}
\end{figure}

Considering the closed-world self-supervised methods that we compared to used different image backbones, traninig epochs and resolutions, we show the corresponding ablation study results in~\tabref{tab:image_bb_ep_reso_ablation}.
As can be seen, the image backbone has the greatest impact on performance. 
However, even using only ResNet-50, our AGO still outperforms all existing methods with 15.23 mIoU on the self-supervised benchmark.
In contrast, the training epochs and the resolution of the input images have relatively small influence.
But even under the most challenging setting, our method is still at the same level as the current state-of-the-art model (with only a 0.36 mIoU gap).
This further indicates that the effectiveness of our method does not come from large numbers of parameters, long training durations, or high-resolution input images, but from the framework design itself.

\Cref{tab:align_loss_ablation} illustrates the impact of different alignment loss functions on the open-world performance of the model.
It can be observed that replacing the cosine similarity loss with either mean squared error (MSE) or mean absolute error (MAE) loss leads to a degradation in prediction performance across all stages, regardless of whether the objects are from known or unknown categories.
Notably, in both the pretraining and zero-shot phases, the model almost entirely loses its ability to recognize unknown instances.
This finding underscores that, in contrast to cosine similarity loss, MSE and MAE losses are not suitable for cross-modal alignment tasks, thereby impairing the perception capability of open-world scenes.

As shown in the~\Cref{tab:mlp_layer_ablation}, we further compare the impact of the number of MLP layers in the modality adapter on the open-world prediction capability.
Notably, when the adapter consists of only a single layer MLP, it does not include any non-linear activation functions.
In this case, the semantic space before and after adaptation remains highly similar, leading to performance comparable to Gro.+Align in~\Cref{tab:alignment_ablation}.
As the number of non-linear projection layers increases, a clear trend is observed: while the mIoU for unknown categories has a slight improvement, the mIoU for known categories degrades significantly.
Considering the overall predictive performance, we use a two-layer MLP as the modality adapter.

\subsection{Confidence analysis of 3D embeddings}
\label{subsec:confi_analysis}

In addition to the information entropy of the predicted probability distribution, we also analyze the maximum confidence score of each category during the pretraining phase, \ie the maximum value of the probability distribution..
As shown in the~\Cref{fig:max_confi}, the adaptive 3D embedding exhibits generally higher maximum confidence for unknown categories, while demonstrating more confident predictions for known categories.
This observation aligns with our previous entropy-based analysis in~\Cref{subsec:ablation_study}, where predictions with lower entropy tend to correspond to higher confidence scores.
Therefore, maximum confidence can also serve as a criterion in the open world identifier.

\subsection{Comparison under VEON's open-world setting}
\label{subsec:veon_comparison}

\begin{table}[t]
    \centering
    \footnotesize
    \resizebox{0.9\linewidth}{!}{
        \begin{tabular}{c|c|ccc}
            \toprule
            \multirow{2}{*}{Method} & \multirow{2}{*}{Param.} & \multicolumn{3}{c}{Seen/Unseen mIoU} \\
            & & 0/17 & 9/8 & 13/4 \\
            \midrule  
            VEON~\cite{zheng2024veon} & 678.1M & 15.14 & 15.16 & 19.94 \\
            AGO  & 62.5M  & \textbf{19.23} & \textbf{22.42} & \textbf{25.90} \\ 
            \bottomrule
        \end{tabular}
    }
    \vspace{-5pt}
    \caption{
    \textbf{VEON's open-world benchmark.}
    }
    \label{tab:veon_comparison}
    \vspace{-5pt}
\end{table}

VEON~\cite{zheng2024veon} defines another open-world benchmark with partial semantic labels.
Specifically, it divides the complete label set into X seen categories with GT annotations and Y unseen categories without annotations for training, and then performs inference and evaluation on the complete label set.
In~\Cref{tab:veon_comparison}, we also compare AGO with it under the same settings.
It can be observed that, regardless of the X/Y setting, our method consistently outperforms VEON~\cite{zheng2024veon} by a significant margin, while utilizing less than 10\% of its parameters.

\subsection{Comparison on Occ3D-Waymo dataset}
\label{subsec:selfsupervised_waymo_benchmark}
In~\Cref{tab:selfsupervised_waymo_benchmark}, we provide the closed-world comparison based on the Occ3D-Waymo~\cite{tian2024occ3d} dataset.
In addition, \Cref{tab:open_world_waymo_benchmark} presents the prediction performance comparison under the open-world setting, with the ``GO'' category excluded due to its semantic ambiguity.

\begin{table*}[htbp]
    \vspace{-5pt}
    \centering
    \resizebox{\textwidth}{!}{
    \begin{tabular}{l|c|ccccccccccccccc|c}
        \toprule
        Method 
        & Image Backbone
        & \rotatebox{90}{\parbox{2cm}{\raggedright\tikz\draw[fill={rgb,255: red,0; green,0; blue,0}, draw={rgb,255: red,0; green,0; blue,0}] (0,0) rectangle (0.2,0.2);~GO}} 
        & \rotatebox{90}{\parbox{2cm}{\raggedright\tikz\draw[fill={rgb,255: red,220; green,20; blue,60}, draw={rgb,255: red,220; green,20; blue,60}] (0,0) rectangle (0.2,0.2);~vehicle}} 
        & \rotatebox{90}{\parbox{2cm}{\raggedright\tikz\draw[fill={rgb,255: red,0; green,0; blue,230}, draw={rgb,255: red,0; green,0; blue,230}] (0,0) rectangle (0.2,0.2);~bicyclist}} 
        & \rotatebox{90}{\parbox{2cm}{\raggedright\tikz\draw[fill={rgb,255: red,47; green,79; blue,79}, draw={rgb,255: red,47; green,79; blue,79}] (0,0) rectangle (0.2,0.2);~pedestrian}}
        & \rotatebox{90}{\parbox{2cm}{\raggedright\tikz\draw[fill={rgb,255: red,0; green,150; blue,245}, draw={rgb,255: red,0; green,150; blue,245}] (0,0) rectangle (0.2,0.2);~sign}} 
        & \rotatebox{90}{\parbox{2cm}{\raggedright\tikz\draw[fill={rgb,255: red,255; green,69; blue,0}, draw={rgb,255: red,255; green,69; blue,0}] (0,0) rectangle (0.2,0.2);~traffic light}}
        & \rotatebox{90}{\parbox{2cm}{\raggedright\tikz\draw[fill={rgb,255: red,255; green,140; blue,0}, draw={rgb,255: red,255; green,140; blue,0}] (0,0) rectangle (0.2,0.2);~pole}} 
        & \rotatebox{90}{\parbox{2cm}{\raggedright\tikz\draw[fill={rgb,255: red,233; green,150; blue,70}, draw={rgb,255: red,233; green,150; blue,70}] (0,0) rectangle (0.2,0.2);~cons. cone}}
        & \rotatebox{90}{\parbox{2cm}{\raggedright\tikz\draw[fill={rgb,255: red,255; green,61; blue,99}, draw={rgb,255: red,255; green,61; blue,99}] (0,0) rectangle (0.2,0.2);~bicycle}}
        & \rotatebox{90}{\parbox{2cm}{\raggedright\tikz\draw[fill={rgb,255: red,112; green,128; blue,144}, draw={rgb,255: red,112; green,128; blue,144}] (0,0) rectangle (0.2,0.2);~motorcycle}} 
        & \rotatebox{90}{\parbox{2cm}{\raggedright\tikz\draw[fill={rgb,255: red,222; green,184; blue,135}, draw={rgb,255: red,222; green,184; blue,135}] (0,0) rectangle (0.2,0.2);~building}}
        & \rotatebox{90}{\parbox{2cm}{\raggedright\tikz\draw[fill={rgb,255: red,0; green,175; blue,0}, draw={rgb,255: red,0; green,175; blue,0}] (0,0) rectangle (0.2,0.2);~vegetation}}
        & \rotatebox{90}{\parbox{2cm}{\raggedright\tikz\draw[fill={rgb,255: red,165; green,42; blue,42}, draw={rgb,255: red,165; green,42; blue,42}] (0,0) rectangle (0.2,0.2);~tree trunk}}
        & \rotatebox{90}{\parbox{2cm}{\raggedright\tikz\draw[fill={rgb,255: red,0; green,207; blue,191}, draw={rgb,255: red,0; green,207; blue,191}] (0,0) rectangle (0.2,0.2);~road}}
        & \rotatebox{90}{\parbox{2cm}{\raggedright\tikz\draw[fill={rgb,255: red,75; green,0; blue,75}, draw={rgb,255: red,75; green,0; blue,75}] (0,0) rectangle (0.2,0.2);~sidewalk}}
        & \rotatebox{90}{mIoU}\\
        \midrule            
        POP-3D$^{\dagger}$~\cite{vobecky2024pop} & ResNet-101 & 0.31 & 20.31 & 5.46 & \textbf{0.83} & 0.00 & 7.11& 10.02 & 7.29 & 9.36 & 0.65 & 12.62 & 8.90 & 1.60 & 65.51 & 18.89 & 11.26 \\
        SelfOcc$^{\dagger}$~\cite{huang2024selfocc} & ResNet-50 & 1.06 & 22.90 & 6.38 & 0.52 & 0.00 & 9.03 & 14.88 & 6.68 & 11.25 & 0.23 & 15.10 & 8.81 & 2.96 & 69.77 & \textbf{22.32} & 12.79 \\
        GaussTR$^{\dagger}$~\cite{jiang2024gausstr} & VFMs & \textbf{2.10} & \textbf{23.13} & \textbf{\underline{7.15}} & 0.15 & 0.00 & \textbf{10.71} & \textbf{15.59} & \textbf{8.21} & \textbf{12.86} & \textbf{0.89} & \textbf{19.52} & \textbf{12.25} & \textbf{3.69} & \textbf{70.95} & \textbf{\underline{22.41}} & \textbf{13.97} \\
        \midrule
        AGO (ours) & ResNet-101 & \textbf{\underline{2.49}} & \textbf{\underline{26.97}} & \textbf{6.94} & \textbf{\underline{1.01}} & 0.00 & \textbf{\underline{25.02}} & \textbf{\underline{17.95}} & \textbf{\underline{11.47}} & \textbf{\underline{14.15}} & \textbf{\underline{1.87}} & \textbf{\underline{22.06}} & \textbf{\underline{14.56}} & \textbf{\underline{3.77}} & \textbf{\underline{71.72}} & 20.26 & \textbf{\underline{16.02}} \\
        \bottomrule
    \end{tabular}}
    \vspace{-5pt}
    \caption{
    \textbf{3D occupancy prediction performance under the self-supervised setting on the Occ3D-Waymo~\cite{tian2024occ3d} dataset.} 
    ``cons. cone'' stands for construction cone.
    $^{\dagger}$ indicate values obtained from our retraining. 
    Results are highlighted in \textbf{\underline{bold \& underlined}} for the best performance and \textbf{bold} for the second-best performance.
    }
    \label{tab:selfsupervised_waymo_benchmark}
    \vspace{5pt}
\end{table*}

\begin{table*}[htbp]
    \vspace{-5pt}
    \centering
    \resizebox{\textwidth}{!}{
    \begin{tabular}{l|l|cccc|c|c|cc|cccccccc|c|c}
        \toprule
        Training Stages
        & Method
        & \rotatebox{90}{\parbox{2cm}{\raggedright\tikz\draw[fill={rgb,255: red,47; green,79; blue,79}, draw={rgb,255: red,47; green,79; blue,79}] (0,0) rectangle (0.2,0.2);~pedestrian}}
        & \rotatebox{90}{\parbox{2cm}{\raggedright\tikz\draw[fill={rgb,255: red,0; green,207; blue,191}, draw={rgb,255: red,0; green,207; blue,191}] (0,0) rectangle (0.2,0.2);~road}}
        & \rotatebox{90}{\parbox{2cm}{\raggedright\tikz\draw[fill={rgb,255: red,75; green,0; blue,75}, draw={rgb,255: red,75; green,0; blue,75}] (0,0) rectangle (0.2,0.2);~sidewalk}}
        & \rotatebox{90}{\parbox{2cm}{\raggedright\tikz\draw[fill={rgb,255: red,220; green,20; blue,60}, draw={rgb,255: red,220; green,20; blue,60}] (0,0) rectangle (0.2,0.2);~vehicle}} 
        & \rotatebox{90}{\parbox{2cm}{\raggedright\tikz\draw[fill={rgb,255: red,128; green,0; blue,128}, draw={rgb,255: red,128; green,0; blue,128}] (0,0) rectangle (0.2,0.2);~cycle}}
        & \rotatebox{90}{k. mIoU}
        & \rotatebox{90}{\parbox{2cm}{\raggedright\tikz\draw[fill={rgb,255: red,255; green,61; blue,99}, draw={rgb,255: red,255; green,61; blue,99}] (0,0) rectangle (0.2,0.2);~bicycle}}
        & \rotatebox{90}{\parbox{2cm}{\raggedright\tikz\draw[fill={rgb,255: red,112; green,128; blue,144}, draw={rgb,255: red,112; green,128; blue,144}] (0,0) rectangle (0.2,0.2);~motorcycle}} 
        & \rotatebox{90}{\parbox{2cm}{\raggedright\tikz\draw[fill={rgb,255: red,0; green,0; blue,230}, draw={rgb,255: red,0; green,0; blue,230}] (0,0) rectangle (0.2,0.2);~bicyclist}} 
        & \rotatebox{90}{\parbox{2cm}{\raggedright\tikz\draw[fill={rgb,255: red,0; green,150; blue,245}, draw={rgb,255: red,0; green,150; blue,245}] (0,0) rectangle (0.2,0.2);~sign}} 
        & \rotatebox{90}{\parbox{2cm}{\raggedright\tikz\draw[fill={rgb,255: red,255; green,69; blue,0}, draw={rgb,255: red,255; green,69; blue,0}] (0,0) rectangle (0.2,0.2);~traffic light}}
        & \rotatebox{90}{\parbox{2cm}{\raggedright\tikz\draw[fill={rgb,255: red,255; green,140; blue,0}, draw={rgb,255: red,255; green,140; blue,0}] (0,0) rectangle (0.2,0.2);~pole}} 
        & \rotatebox{90}{\parbox{2cm}{\raggedright\tikz\draw[fill={rgb,255: red,233; green,150; blue,70}, draw={rgb,255: red,233; green,150; blue,70}] (0,0) rectangle (0.2,0.2);~cons. cone}}
        & \rotatebox{90}{\parbox{2cm}{\raggedright\tikz\draw[fill={rgb,255: red,222; green,184; blue,135}, draw={rgb,255: red,222; green,184; blue,135}] (0,0) rectangle (0.2,0.2);~building}}
        & \rotatebox{90}{\parbox{2cm}{\raggedright\tikz\draw[fill={rgb,255: red,0; green,175; blue,0}, draw={rgb,255: red,0; green,175; blue,0}] (0,0) rectangle (0.2,0.2);~vegetation}}
        & \rotatebox{90}{\parbox{2cm}{\raggedright\tikz\draw[fill={rgb,255: red,165; green,42; blue,42}, draw={rgb,255: red,165; green,42; blue,42}] (0,0) rectangle (0.2,0.2);~tree trunk}}
        & \rotatebox{90}{u. mIoU}
        & \rotatebox{90}{mIoU} \\
        
        \midrule
        
        \multirow{4}{*}{\makecell[l]{Pretraining}} & POP-3D$^{\dagger}$~\cite{vobecky2024pop} & \cellcolor{green!25}0.00 & \cellcolor{green!25}49.84 & \cellcolor{green!25}6.89 & \cellcolor{green!25}10.43 & \cellcolor{cyan!25}0.00 & 13.43 & - & - & \cellcolor{red!25}0.00 & \cellcolor{red!25}0.00 & \cellcolor{red!25}0.00 & \cellcolor{red!25}0.00 & \cellcolor{red!25}0.00 & \cellcolor{red!25}1.26 & \cellcolor{red!25}0.57 & \cellcolor{red!25}0.44 & 0.28 & 5.34 \\
        
        & SelfOcc$^{\dagger}$~\cite{huang2024selfocc} & \cellcolor{green!25}\textbf{0.27} & \cellcolor{green!25}67.45 & \cellcolor{green!25}16.07 & \cellcolor{green!25}28.05 & \cellcolor{cyan!25}0.00 & 22.37 & - & - & \cellcolor{red!25}0.00 & \cellcolor{red!25}0.00 & \cellcolor{red!25}0.00 & \cellcolor{red!25}0.00 & \cellcolor{red!25}0.00 & \cellcolor{red!25}0.00 & \cellcolor{red!25}0.00 & \cellcolor{red!25}0.00 & 0.00 & 8.60 \\

        & GaussTR$^{\dagger}$~\cite{jiang2024gausstr} & \cellcolor{green!25}0.22 & \cellcolor{green!25}\textbf{69.97} & \cellcolor{green!25}\textbf{22.85} & \cellcolor{green!25}12.05 & \cellcolor{cyan!25}0.00 & 21.02 & - & - & \cellcolor{red!25}0.00 & \cellcolor{red!25}\textbf{0.04} & \cellcolor{red!25}0.00 & \cellcolor{red!25}0.00 & \cellcolor{red!25}0.00 & \cellcolor{red!25}4.20 & \cellcolor{red!25}0.99 & \cellcolor{red!25}0.29 & 0.69 & 8.51 \\
        
        & AGO (ours) & \cellcolor{green!25}0.09 & \cellcolor{green!25}67.59 & \cellcolor{green!25}19.50 & \cellcolor{green!25}\textbf{28.29} & \cellcolor{cyan!25}\textbf{6.92} & \textbf{24.48} & - & - & \cellcolor{red!25}\textbf{0.03} & \cellcolor{red!25}0.00 & \cellcolor{red!25}\textbf{0.01} & \cellcolor{red!25}0.00 & \cellcolor{red!25}\textbf{0.05} & \cellcolor{red!25}\textbf{5.16} & \cellcolor{red!25}\textbf{1.92} & \cellcolor{red!25}\textbf{0.61} & \textbf{0.97} & \textbf{10.01} \\
        
        \midrule
        
        \multirow{4}{*}{\makecell[l]{Zero-shot \\ Evaluation}} & POP-3D$^{\dagger}$~\cite{vobecky2024pop} & \cellcolor{green!25}0.00 & \cellcolor{green!25}49.84 & \cellcolor{green!25}6.89 & \cellcolor{green!25}10.43 & - & 16.79 & \cellcolor{orange!25}0.00 & \cellcolor{orange!25}0.00 & \cellcolor{red!25}0.00 & \cellcolor{red!25}0.00 & \cellcolor{red!25}0.00 & \cellcolor{red!25}0.00 & \cellcolor{red!25}0.00 & \cellcolor{red!25}1.26 & \cellcolor{red!25}0.57 & \cellcolor{red!25}0.44 & 0.23 & 4.96\\
        
        & SelfOcc$^{\dagger}$~\cite{huang2024selfocc} & \cellcolor{green!25}\textbf{0.27} & \cellcolor{green!25}67.45 & \cellcolor{green!25}16.07 & \cellcolor{green!25}28.05 & - & 27.96 & \cellcolor{orange!25}0.00 & \cellcolor{orange!25}0.00 & \cellcolor{red!25}0.00 & \cellcolor{red!25}0.00 & \cellcolor{red!25}0.00 & \cellcolor{red!25}0.00 & \cellcolor{red!25}0.00 & \cellcolor{red!25}0.00 & \cellcolor{red!25}0.00 & \cellcolor{red!25}0.00 & 0.00 & 7.99\\

        & GaussTR$^{\dagger}$~\cite{jiang2024gausstr} & \cellcolor{green!25}0.22 & \cellcolor{green!25}\textbf{69.97} & \cellcolor{green!25}\textbf{22.85} & \cellcolor{green!25}12.05 & - & 26.27 & \cellcolor{orange!25}0.00 & \cellcolor{orange!25}0.00 & \cellcolor{red!25}0.00 & \cellcolor{red!25}\textbf{0.04} & \cellcolor{red!25}0.00 & \cellcolor{red!25}0.00 & \cellcolor{red!25}0.00 & \cellcolor{red!25}4.20 & \cellcolor{red!25}0.99 & \cellcolor{red!25}0.29 & 0.55 & 7.90\\
        
        & AGO (ours) & \cellcolor{green!25}0.09 & \cellcolor{green!25}67.59 & \cellcolor{green!25}19.50 & \cellcolor{green!25}\textbf{28.29} & - & \textbf{28.87} & \cellcolor{orange!25}\textbf{5.56} & \cellcolor{orange!25}\textbf{0.14} & \cellcolor{red!25}\textbf{0.03} & \cellcolor{red!25}0.00 & \cellcolor{red!25}\textbf{0.01} & \cellcolor{red!25}0.00 & \cellcolor{red!25}\textbf{0.05} & \cellcolor{red!25}\textbf{5.16} & \cellcolor{red!25}\textbf{1.92} & \cellcolor{red!25}\textbf{0.61} & \textbf{1.35} & \textbf{9.21}\\
        
        \midrule
        
        \multirow{4}{*}{\makecell[l]{Few-shot \\ Finetuning}} & POP-3D$^{\dagger}$~\cite{vobecky2024pop} & \cellcolor{green!25}0.00 & \cellcolor{green!25}37.96 & \cellcolor{green!25}6.26 & \cellcolor{green!25}9.12 & - & 13.34 & \cellcolor{orange!25}0.00 & \cellcolor{orange!25}0.00 & \cellcolor{red!25}0.00 & \cellcolor{red!25}0.00 & \cellcolor{red!25}0.00 & \cellcolor{red!25}0.00 & \cellcolor{red!25}0.00 & \cellcolor{red!25}1.33 & \cellcolor{red!25}0.49 & \cellcolor{red!25}0.66 & 0.25 & 3.99\\
        
        & SelfOcc$^{\dagger}$~\cite{huang2024selfocc} & \cellcolor{green!25}0.55 & \cellcolor{green!25}68.21 & \cellcolor{green!25}18.20 & \cellcolor{green!25}29.60 & - & 29.14 & \cellcolor{orange!25}1.09 & \cellcolor{orange!25}0.00 & \cellcolor{red!25}0.08 & \cellcolor{red!25}0.00 & \cellcolor{red!25}0.91 & \cellcolor{red!25}0.00 & \cellcolor{red!25}0.00 & \cellcolor{red!25}3.11 & \cellcolor{red!25}1.02 & \cellcolor{red!25}0.03 & 0.62 & 8.77\\

        & GaussTR$^{\dagger}$~\cite{jiang2024gausstr} & \cellcolor{green!25}0.61 & \cellcolor{green!25}69.71 & \cellcolor{green!25}23.10 & \cellcolor{green!25}12.05 & - & 26.37 & \cellcolor{orange!25}5.15 & \cellcolor{orange!25}0.02 & \cellcolor{red!25}2.12 & \cellcolor{red!25}0.00 & \cellcolor{red!25}2.06 & \cellcolor{red!25}1.03 & \cellcolor{red!25}1.27 & \cellcolor{red!25}6.84 & \cellcolor{red!25}3.15 & \cellcolor{red!25}1.12 & 2.28 & 9.16\\
        
        & AGO (ours) & \cellcolor{green!25}\textbf{0.82} & \cellcolor{green!25}\textbf{70.15} & \cellcolor{green!25}\textbf{23.50} & \cellcolor{green!25}\textbf{30.03} & - & \textbf{31.13} & \cellcolor{orange!25}\textbf{9.26} & \cellcolor{orange!25}\textbf{0.56} & \cellcolor{red!25}\textbf{2.13} & \cellcolor{red!25}0.00 & \cellcolor{red!25}\textbf{10.86} & \cellcolor{red!25}\textbf{5.28} & \cellcolor{red!25}\textbf{4.15} & \cellcolor{red!25}\textbf{11.23} & \cellcolor{red!25}\textbf{8.81} & \cellcolor{red!25}\textbf{2.57} & \textbf{5.49} & \textbf{12.81}\\
        \bottomrule
    \end{tabular}}
    \vspace{-5pt}
    \caption{
    \textbf{3D occupancy prediction performance under the open-world setting on the Occ3D-Waymo~\cite{tian2024occ3d} dataset.}
    $^{\dagger}$ indicate values obtained from our retraining.
    The background color represents whether the category is known or unknown during the \textbf{pre-training stage}: \colorbox{green!25}{green} and \colorbox{cyan!25}{blue} indicate known IoU, \colorbox{orange!25}{orange} and \colorbox{red!25}{red} indicate unknown IoU.
    The \colorbox{orange!25}{orange} categories are the refined version of the \colorbox{cyan!25}{blue} categories.
    Results are highlighted in \textbf{bold} for the best performance.
    }
    \label{tab:open_world_waymo_benchmark}
    \vspace{-5pt}
\end{table*}

\subsection{Detailed open-world results}
\label{subsec:detailed_ow_results}
In~\Cref{tab:detailed_open_world_exp}, we provide the detailed prediction results of the open-world ablation experiments in Table~\ref{tab:alignment_ablation} and \ref{tab:ow_strategy_ablation}.

\begin{figure}[t!]
    \centering
    \includegraphics[width=\linewidth]{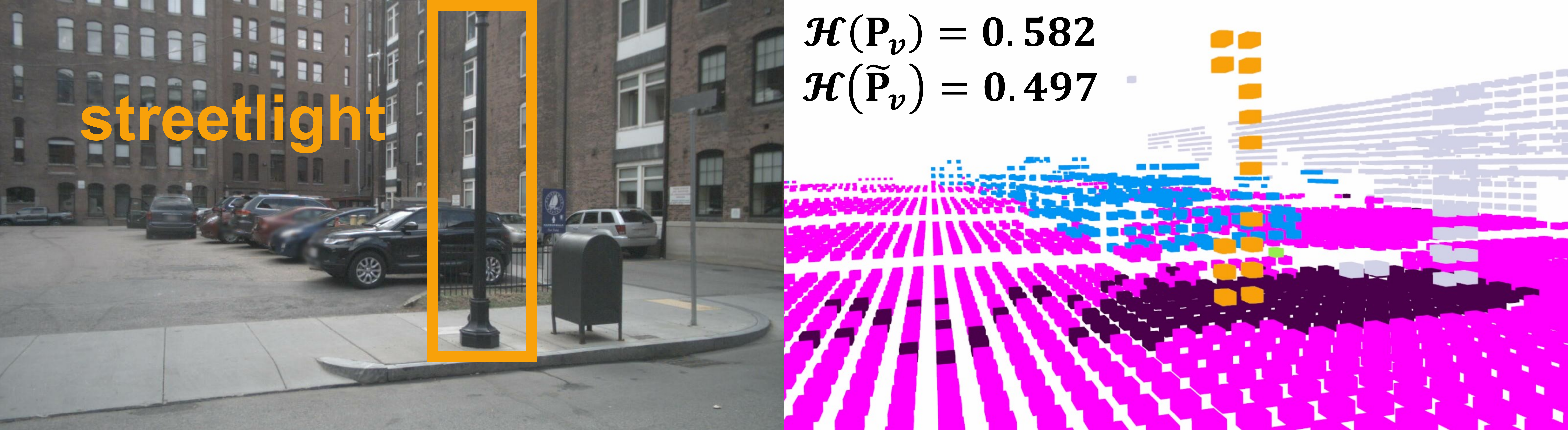}
    \vspace{-15pt}
    \caption{
    \textbf{Visualization of open-vocabular retrieval.} 
    }
    \label{fig:retrieval_vis}
    \vspace{-10pt}
\end{figure}

\section{Additional Visualization}
\label{sec:addl_vis}
\begin{figure*}[htbp]
    \centering
    \includegraphics[width=\textwidth]{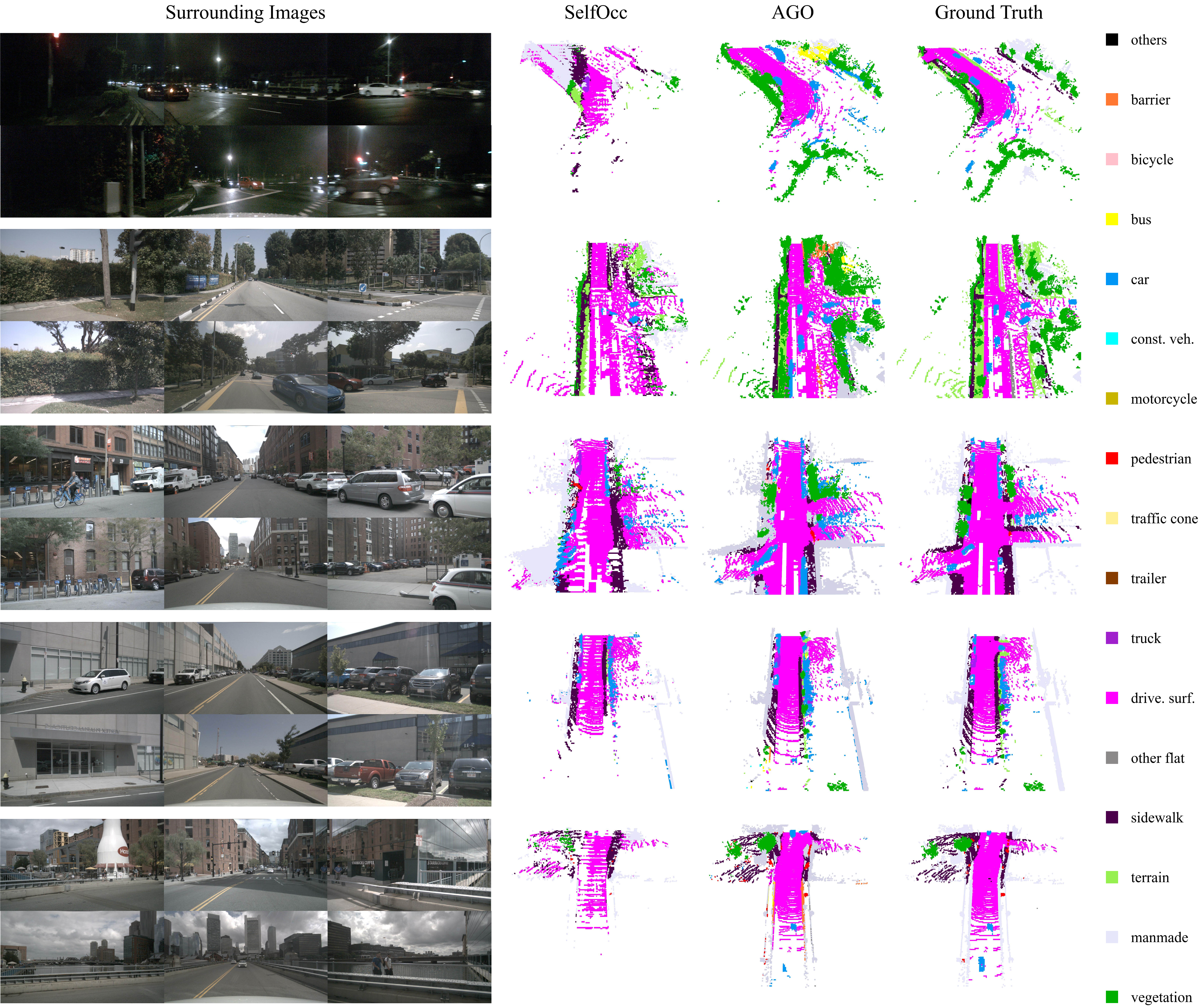}
    \vspace{-20pt}
    \caption{
    \textbf{Additional visualization of self-supervised 3D semantic occupancy prediction on the Occ3D-nuScenes occupancy benchmark.} 
    }
    \label{fig:addl_occ_vis}
\end{figure*}

\Cref{fig:retrieval_vis} shows the visualization of AGO's open vocabulary retrieval results, where $\mathcal{H}(P_v)$ and $\mathcal{H}(\tilde{P}_v)$ represent the average entropy of the corresponding voxel predictions before and after the modality adapter.
We also show more visual comparisons of self-supervised 3D semantic occupancy prediction in~\Cref{fig:addl_occ_vis}.
It can be seen that compared to SelfOcc~\cite{huang2024selfocc}, our AGO provides more complete and fine-grained predictions.
Especially for dynamic categories with small scales, such as cars and pedestrians, the results of our method are closer to the ground truth.
However, due to the natural flaws of volume rendering-based methods in dynamic objects, SelfOcc~\cite{huang2024selfocc} has massive false positives and false negatives in the predictions of these categories.
In addition, limited by the computational complexity, 3D features can only be sampled along the rays at a relatively low sampling rate in volume rendering, which results in a large number of periodic blank strip textures in the final prediction.
This phenomenon is highly evident in the ``driveable surface'' predictions shown in the third and fifth rows of the figure.
Even with constraints such as Hessian loss $\mathcal{L}_H$, regularization $\mathcal{L}_S$, and Eikonal term $\mathcal{L}_E$~\cite{huang2024selfocc}, this issue cannot be fundamentally solved.
These qualitative comparisons further indicate that existing self-supervised models are insufficient for 3D scene understanding.
In contrast, our proposed AGO framework based on grounded training has greater potential in this regard.

\begin{table*}[t]
    \vspace{-5pt}
    \centering
    \resizebox{\textwidth}{!}{
    \begin{tabular}{l|l|ccc|cc|c|ccccccc|ccccc|c|c}
        \toprule
        Training Stages
        & Method
        & \rotatebox{90}{\parbox{2cm}{\raggedright\tikz\draw[fill={rgb,255: red,255; green,0; blue,0}, draw={rgb,255: red,255; green,0; blue,0}] (0,0) rectangle (0.2,0.2);~pedestrian}}
        & \rotatebox{90}{\raggedright\tikz\draw[fill={rgb,255: red,255; green,0; blue,255}, draw={rgb,255: red,255; green,0; blue,255}] (0,0) rectangle (0.2,0.2);~\parbox{2cm}{driveable\\surface}}
        & \rotatebox{90}{\parbox{2cm}{\raggedright\tikz\draw[fill={rgb,255: red,75; green,0; blue,75}, draw={rgb,255: red,75; green,0; blue,75}] (0,0) rectangle (0.2,0.2);~sidewalk}}
        & \rotatebox{90}{\parbox{2cm}{\raggedright\tikz\draw[fill={rgb,255: red,0; green,128; blue,128}, draw={rgb,255: red,0; green,128; blue,128}] (0,0) rectangle (0.2,0.2);~vehicle}}
        & \rotatebox{90}{\parbox{2cm}{\raggedright\tikz\draw[fill={rgb,255: red,128; green,0; blue,128}, draw={rgb,255: red,128; green,0; blue,128}] (0,0) rectangle (0.2,0.2);~cycle}}
        & \rotatebox{90}{known mIoU}
        & \rotatebox{90}{\parbox{2cm}{\raggedright\tikz\draw[fill={rgb,255: red,0; green,150; blue,245}, draw={rgb,255: red,0; green,150; blue,245}] (0,0) rectangle (0.2,0.2);~car}}
        & \rotatebox{90}{\parbox{2cm}{\raggedright\tikz\draw[fill={rgb,255: red,255; green,255; blue,0}, draw={rgb,255: red,255; green,255; blue,0}] (0,0) rectangle (0.2,0.2);~bus}}
        & \rotatebox{90}{\raggedright\tikz\draw[fill={rgb,255: red,0; green,255; blue,255}, draw={rgb,255: red,0; green,255; blue,255}] (0,0) rectangle (0.2,0.2);~\parbox{2cm}{construction\\vehicle}}
        & \rotatebox{90}{\parbox{2cm}{\raggedright\tikz\draw[fill={rgb,255: red,135; green,60; blue,0}, draw={rgb,255: red,135; green,60; blue,0}] (0,0) rectangle (0.2,0.2);~trailer}} 
        & \rotatebox{90}{\parbox{2cm}{\raggedright\tikz\draw[fill={rgb,255: red,160; green,32; blue,240}, draw={rgb,255: red,160; green,32; blue,240}] (0,0) rectangle (0.2,0.2);~truck}}
        & \rotatebox{90}{\parbox{2cm}{\raggedright\tikz\draw[fill={rgb,255: red,255; green,192; blue,203}, draw={rgb,255: red,255; green,192; blue,203}] (0,0) rectangle (0.2,0.2);~bicycle}} 
        & \rotatebox{90}{\parbox{2cm}{\raggedright\tikz\draw[fill={rgb,255: red,200; green,180; blue,0}, draw={rgb,255: red,200; green,180; blue,0}] (0,0) rectangle (0.2,0.2);~motorcycle}} 
        & \rotatebox{90}{\parbox{2cm}{\raggedright\tikz\draw[fill={rgb,255: red,255; green,120; blue,50}, draw={rgb,255: red,255; green,120; blue,50}] (0,0) rectangle (0.2,0.2);~barrier}}  
        & \rotatebox{90}{\parbox{2cm}{\raggedright\tikz\draw[fill={rgb,255: red,255; green,240; blue,150}, draw={rgb,255: red,255; green,240; blue,150}] (0,0) rectangle (0.2,0.2);~traffic cone}}
        & \rotatebox{90}{\parbox{2cm}{\raggedright\tikz\draw[fill={rgb,255: red,150; green,240; blue,80}, draw={rgb,255: red,150; green,240; blue,80}] (0,0) rectangle (0.2,0.2);~terrain}}
        & \rotatebox{90}{\parbox{2cm}{\raggedright\tikz\draw[fill={rgb,255: red,230; green,230; blue,250}, draw={rgb,255: red,230; green,230; blue,250}] (0,0) rectangle (0.2,0.2);~manmade}}
        & \rotatebox{90}{\parbox{2cm}{\raggedright\tikz\draw[fill={rgb,255: red,0; green,175; blue,0}, draw={rgb,255: red,0; green,175; blue,0}] (0,0) rectangle (0.2,0.2);~vegetation}}
        & \rotatebox{90}{unknown mIoU}
        & \rotatebox{90}{mIoU} \\
        
        \midrule
        
        \multirow{5}{*}{\makecell[l]{Pretraining}} & Align & \cellcolor{green!25}0.00 & \cellcolor{green!25}58.10 & \cellcolor{green!25}12.29 & \cellcolor{cyan!25}6.72 & \cellcolor{cyan!25}0.00 & 15.42 & - & - & - & - & - & - & - & \cellcolor{red!25}0.00 & \cellcolor{red!25}0.00 & \cellcolor{red!25}3.81 & \cellcolor{red!25}0.00 & \cellcolor{red!25}0.16 & 0.79 & 8.11 \\
        
        & Gro. & \cellcolor{green!25}1.84 & \cellcolor{green!25}68.68 & \cellcolor{green!25}28.52 & \cellcolor{cyan!25}3.15 & \cellcolor{cyan!25}0.99 & 20.64 & - & - & - & - & - & - & - & \cellcolor{red!25}0.00 & \cellcolor{red!25}0.00 & \cellcolor{red!25}0.00 & \cellcolor{red!25}0.00 & \cellcolor{red!25}0.00 & 0.00 & 10.32 \\

        & Gro. + Align & \cellcolor{green!25}4.31 & \cellcolor{green!25}57.73 & \cellcolor{green!25}25.96 & \cellcolor{cyan!25}2.36 & \cellcolor{cyan!25}0.90 & 18.25 & - & - & - & - & - & - & - & \cellcolor{red!25}0.00 & \cellcolor{red!25}0.00 & \cellcolor{red!25}2.42 & \cellcolor{red!25}0.02 & \cellcolor{red!25}8.49 & 2.19 & 10.22 \\
        
        & AGO w/o OWI & \cellcolor{green!25}7.25 & \cellcolor{green!25}65.27 & \cellcolor{green!25}25.93 & \cellcolor{cyan!25}7.82 & \cellcolor{cyan!25}4.92 & 22.24 & - & - & - & - & - & - & - & \cellcolor{red!25}0.00 & \cellcolor{red!25}0.00 & \cellcolor{red!25}0.78 & \cellcolor{red!25}0.00 & \cellcolor{red!25}4.63 & 1.08 & 11.66 \\

        & AGO w/ Max Confi. & \cellcolor{green!25}7.36 & \cellcolor{green!25}64.76 & \cellcolor{green!25}25.68 & \cellcolor{cyan!25}8.92 & \cellcolor{cyan!25}5.32 & 22.41 & - & - & - & - & - & - & - & \cellcolor{red!25}0.00 & \cellcolor{red!25}0.00 & \cellcolor{red!25}6.90 & \cellcolor{red!25}0.01 & \cellcolor{red!25}8.73 & 3.13 & 12.77 \\
        
        \midrule
        
        \multirow{5}{*}{\makecell[l]{Zero-shot \\ Evaluation}} & Align & \cellcolor{green!25}0.00 & \cellcolor{green!25}58.10 & \cellcolor{green!25}12.29 & - & - & 23.46 & \cellcolor{orange!25}6.81 & \cellcolor{orange!25}0.00 & \cellcolor{orange!25}0.00 & \cellcolor{orange!25}0.57 & \cellcolor{orange!25}2.91 & \cellcolor{orange!25}0.00 & \cellcolor{orange!25}0.00 & \cellcolor{red!25}0.00 & \cellcolor{red!25}0.00 & \cellcolor{red!25}3.81 & \cellcolor{red!25}0.00 & \cellcolor{red!25}0.16 & 1.19 & 5.64 \\
        
        & Gro. & \cellcolor{green!25}1.84 & \cellcolor{green!25}68.68 & \cellcolor{green!25}28.52 & - & - & 33.01 & \cellcolor{orange!25}2.21 & \cellcolor{orange!25}0.00 & \cellcolor{orange!25}0.00 & \cellcolor{orange!25}0.02 & \cellcolor{orange!25}0.87 & \cellcolor{orange!25}0.00 & \cellcolor{orange!25}0.00 & \cellcolor{red!25}0.00 & \cellcolor{red!25}0.00 & \cellcolor{red!25}0.00 & \cellcolor{red!25}0.00 & \cellcolor{red!25}0.00 & 0.26 & 6.81\\

        & Gro. + Align & \cellcolor{green!25}4.31 & \cellcolor{green!25}57.73 & \cellcolor{green!25}25.96 & - & - & 29.33 & \cellcolor{orange!25}2.78 & \cellcolor{orange!25}0.00 & \cellcolor{orange!25}0.00 & \cellcolor{orange!25}0.15 & \cellcolor{orange!25}1.29 & \cellcolor{orange!25}1.36 & \cellcolor{orange!25}0.00 & \cellcolor{red!25}0.00 & \cellcolor{red!25}0.00 & \cellcolor{red!25}2.42 & \cellcolor{red!25}0.02 & \cellcolor{red!25}8.49 & 1.38 & 6.97\\

        & AGO w/o OWI & \cellcolor{green!25}7.25 & \cellcolor{green!25}65.27 & \cellcolor{green!25}25.93 & - & - & 32.82 & \cellcolor{orange!25}5.26 & \cellcolor{orange!25}0.00 & \cellcolor{orange!25}0.00 & \cellcolor{orange!25}0.82 & \cellcolor{orange!25}4.36 & \cellcolor{orange!25}3.08 & \cellcolor{orange!25}0.00 & \cellcolor{red!25}0.00 & \cellcolor{red!25}0.00 & \cellcolor{red!25}0.78 & \cellcolor{red!25}0.00 & \cellcolor{red!25}4.63 & 1.58 & 7.83\\
        
        & AGO w/ Max Confi. & \cellcolor{green!25}7.36 & \cellcolor{green!25}64.76 & \cellcolor{green!25}25.68 & - & - & 32.60 & \cellcolor{orange!25}7.63 & \cellcolor{orange!25}0.00 & \cellcolor{orange!25}0.00 & \cellcolor{orange!25}0.71 & \cellcolor{orange!25}6.08 & \cellcolor{orange!25}3.13 & \cellcolor{orange!25}0.00 & \cellcolor{red!25}0.00 & \cellcolor{red!25}0.00 & \cellcolor{red!25}6.90 & \cellcolor{red!25}0.01 & \cellcolor{red!25}8.73 & 2.77 & 8.73\\
        
        \midrule
        
        \multirow{5}{*}{\makecell[l]{Few-shot \\ Finetuning}} & Align & \cellcolor{green!25}0.00 & \cellcolor{green!25}58.95 & \cellcolor{green!25}14.18 & - & - & 24.38 & \cellcolor{orange!25}5.64 & \cellcolor{orange!25}0.00 & \cellcolor{orange!25}0.00 & \cellcolor{orange!25}0.50 & \cellcolor{orange!25}3.97 & \cellcolor{orange!25}0.00 & \cellcolor{orange!25}0.00 & \cellcolor{red!25}0.00 & \cellcolor{red!25}0.00 & \cellcolor{red!25}11.70 & \cellcolor{red!25}11.34 & \cellcolor{red!25}15.28 & 4.04 & 8.10\\
        
        & Gro. & \cellcolor{green!25}13.34 & \cellcolor{green!25}70.95 & \cellcolor{green!25}30.90 & - & - & 38.40 & \cellcolor{orange!25}15.94 & \cellcolor{orange!25}0.00 & \cellcolor{orange!25}0.00 & \cellcolor{orange!25}0.00 & \cellcolor{orange!25}0.00 & \cellcolor{orange!25}0.00 & \cellcolor{orange!25}0.00 & \cellcolor{red!25}0.00 & \cellcolor{red!25}0.00 & \cellcolor{red!25}0.02 & \cellcolor{red!25}8.14 & \cellcolor{red!25}12.25 & 3.03 & 10.10\\

        & Gro. + Align & \cellcolor{green!25}12.01 & \cellcolor{green!25}71.57 & \cellcolor{green!25}28.44 & - & - & 37.34 & \cellcolor{orange!25}13.52 & \cellcolor{orange!25}4.11 & \cellcolor{orange!25}0.00 & \cellcolor{orange!25}0.00 & \cellcolor{orange!25}1.36 & \cellcolor{orange!25}1.24 & \cellcolor{orange!25}0.00 & \cellcolor{red!25}0.00 & \cellcolor{red!25}0.00 & \cellcolor{red!25}26.83 & \cellcolor{red!25}8.86 & \cellcolor{red!25}12.68 & 5.72 & 12.04\\

        & AGO w/o OWI & \cellcolor{green!25}12.58 & \cellcolor{green!25}72.02 & \cellcolor{green!25}30.12 & - & - & 38.24 & \cellcolor{orange!25}18.69 & \cellcolor{orange!25}5.20 & \cellcolor{orange!25}0.00 & \cellcolor{orange!25}0.21 & \cellcolor{orange!25}2.56 & \cellcolor{orange!25}3.59 & \cellcolor{orange!25}2.18 & \cellcolor{red!25}0.13 & \cellcolor{red!25}0.00 & \cellcolor{red!25}29.39 & \cellcolor{red!25}21.54 & \cellcolor{red!25}17.80 & 8.44 & 14.40\\
        
        & AGO w/ Max Confi. & \cellcolor{green!25}12.97 & \cellcolor{green!25}71.53 & \cellcolor{green!25}29.63 & - & - & 38.04 & \cellcolor{orange!25}18.58 & \cellcolor{orange!25}5.03 & \cellcolor{orange!25}0.00 & \cellcolor{orange!25}0.23 & \cellcolor{orange!25}2.74 & \cellcolor{orange!25}3.56 & \cellcolor{orange!25}2.04 & \cellcolor{red!25}0.30 & \cellcolor{red!25}0.00 & \cellcolor{red!25}29.33 & \cellcolor{red!25}21.32 & \cellcolor{red!25}17.81 & 8.41 & 14.34\\
        \bottomrule
    \end{tabular}}
    \vspace{-8pt}
    \caption{
    \textbf{Detailed 3D occupancy prediction results under the open-world setting on the Occ3D-nuScenes~\cite{tian2024occ3d} dataset.}
    The background color represents whether the category is known or unknown during the \textbf{pre-training stage}: \colorbox{green!25}{green} and \colorbox{cyan!25}{blue} indicate known IoU, \colorbox{orange!25}{orange} and \colorbox{red!25}{red} indicate unknown IoU.
    The \colorbox{orange!25}{orange} categories are the refined version of the \colorbox{cyan!25}{blue} categories.
    }
    \label{tab:detailed_open_world_exp}
    \vspace{-10pt}
\end{table*}

\end{document}